\documentclass{article} 
\usepackage{iclr2027_conference,times}

\usepackage{amsmath,amsfonts,bm}

\def\eqref#1{equation~\ref{#1}}
\def\Eqref#1{Equation~\ref{#1}}

\def\1{\bm{1}}

\def\vc{{\bm{c}}}

\def\vp{{\bm{p}}}

\def\vr{{\bm{r}}}

\def\vx{{\bm{x}}}

\DeclareMathAlphabet{\mathsfit}{\encodingdefault}{\sfdefault}{m}{sl}
\SetMathAlphabet{\mathsfit}{bold}{\encodingdefault}{\sfdefault}{bx}{n}

\def\sA{{\mathbb{A}}}

\def\sM{{\mathbb{M}}}

\def\sU{{\mathbb{U}}}
\def\sV{{\mathbb{V}}}

\usepackage{hyperref}
\usepackage{url}
\usepackage{graphicx}
\usepackage{colortbl}
\usepackage{multirow}
\usepackage{booktabs}
\usepackage{wrapfig}
\usepackage{enumitem}
\usepackage{subcaption}
\usepackage{placeins}
\usepackage{tcolorbox}
\tcbuselibrary{skins}
\usepackage{algorithm}
\usepackage{algorithmic}

\title{FastGuide: Accelerating Reward Guidance for Diffusion Large Language Models}

\author{%
    Darshan Thaker, Lachlan Ewen MacDonald \& Ren\'e Vidal \\
    University of Pennsylvania \\
    \texttt{\{dbthaker,lmacdo,vidalr\}@seas.upenn.edu}
}

\newcommand{\mask}{\texttt{<MASK>}}
\newcommand{\ours}{FastGuide}
\newcommand{\g}{\rowcolor{gray!10}}

\definecolor{obsone}{RGB}{0,109,176}
\definecolor{obstwo}{RGB}{0,109,176}
\newcommand{\obsone}{\textcolor{obsone}{\textbf{O1}}}
\newcommand{\obstwo}{\textcolor{obstwo}{\textbf{O2}}}

\newif\ifrewardfigs
\rewardfigstrue

\iclrfinalcopy 
\begin{document}
\addtocontents{toc}{\protect\setcounter{tocdepth}{-10}}

\maketitle

\begin{abstract}
    Gradient-based reward guidance provides a flexible way to use downstream reward models to control masked diffusion language models at inference time. However, its computational cost remains high as each decoding iteration incurs expensive diffusion model forward passes and reward model backpropagation steps. To address this, we introduce \ours{}, an adaptive hybrid of parallel and autoregressive decoding to accelerate reward guidance for diffusion language models. In analogy to parallel decoding, \ours{} amortizes the cost of reward model backpropagation by computing guidance once per decoding step and reusing it to generate multiple tokens. Within each decoding step, \ours{} makes diffusion forward passes autoregressive by unmasking tokens one at a time while efficiently recomputing token distributions after each unmasking by utilizing KV caching techniques and sparse recomputation of attention. Lastly, to adapt hybrid decoding to the model's confidence, \ours{} defers any token that the model is unconfident about under its recomputed distribution. Experiments on three reward benchmarks demonstrate that \ours{} is up to $4.4\times$ faster than sequential reward-guided decoding while retaining similar generation quality. 
\end{abstract}    

\section{Introduction}
Diffusion Large Language Models (dLLMs) have emerged as a promising alternative to traditional autoregressive models and have been praised for their reasoning power and generation speed \citep{ye2025beyond,nie2026large}. A popular class of dLLMs is discrete masked dLLMs \citep{austin2021structured, sahoo2024simple}, which are trained on partially masked token sequences to predict marginal distributions over the vocabulary for each masked token. At inference time, dLLM generation starts from a fully masked sequence of fixed length and decoding algorithms iteratively decide a) where to unmask, and b) which token values to place at the selected positions. For instance, a common strategy is to compute a confidence score for each token position, pick the most confident token, and sample from the corresponding marginal distribution of that token \citep{ye2025dream}.

As these models begin to be deployed in practical settings \citep{labs2025mercury, geminidiffusion}, an important area of research is ensuring alignment of the dLLM outputs to desired objectives, e.g., instruction following, truthfulness, or safety \citep{wen2026devil,xiong2026unveiling}. A recent line of work has explored training-free gradient-based guidance algorithms for dLLM alignment using a differentiable reward model \citep{murata2024g2d2, tejaswi2026entropy}. For these approaches, at each diffusion timestep $t$ with partially masked sequence $\vx_t$ and unmasked tokens $\vx_t^\text{u}$, the dLLM logits at each token position $i$ are steered via the updates
\begin{equation} \label{eq:guidance}
    \ell_t^{\text{guided}}(x_t^i \mid \vx_t^{\text{u}}) = \underbrace{\ell_t^{\text{unguided}}(x_t^i \mid \vx_t^{\text{u}})}_{\text{base dLLM logits}} + \underbrace{\vr_t(x_t^i \mid \vx_t^{\text{u}})}_{\text{guidance vector}}.
\end{equation}
Above, $\ell_t^{\text{unguided}}$ denotes the per-token dLLM logits conditioned on the unmasked tokens $\vx_t^{\text{u}}$, and $\vr_t$ denotes the corresponding per-token guidance vector. The guidance vector is estimated by cheaply constructing a fully unmasked sequence, evaluating that sequence with a reward model, and computing gradients of the reward model with respect to its input to steer the dLLM logits. Although gradient-based guidance can outperform alternatives such as Best-Of-N sampling \citep{tejaswi2026entropy}, the sampling speed of these approaches is slow, as shown in the top row of Figure \ref{fig:teaser}b. This is due to the dependency structure of \Eqref{eq:guidance}. At each diffusion timestep, when one token is unmasked, both the dLLM logits and the guidance vector change because the conditioning variable changes. Recomputing the dLLM logits requires a forward pass through the dLLM, which involves bidirectional attention over the sequence, and recomputing the guidance vector requires a forward and backward pass through a large reward model (e.g., a Skywork-Llama-8B model \citep{liu2026skywork}). Both of these steps can be large inference-time bottlenecks.

Parallel decoding is the dominant strategy to accelerate dLLM inference, so a natural idea is to follow existing work and unmask multiple tokens in each diffusion timestep \citep{eb_sampler, kim2026dapd, fastdllm}. However, existing work is developed for unguided generation, so it does not establish when guidance can be reused, which is the main inference-time bottleneck (Figure \ref{fig:teaser}b). Further, it is unclear which tokens remain safe to unmask together once guidance has altered the token distributions. This leads to our main research question: 

\begin{figure}[t]
    \centering
    \includegraphics[width=0.87\linewidth]{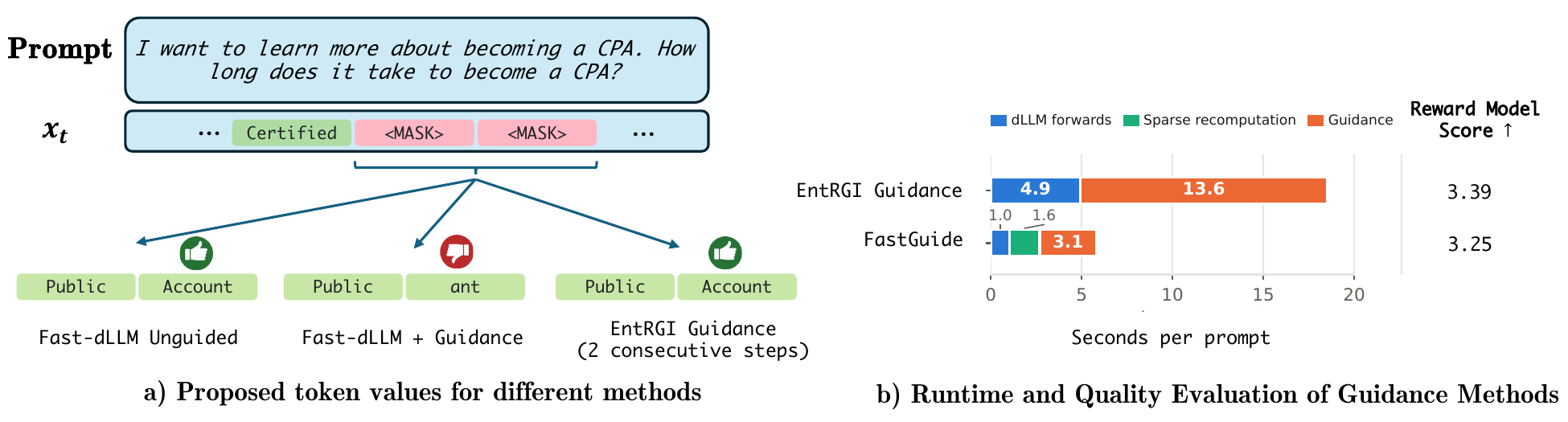}
    \caption{Left: Reward guidance \citep{tejaswi2026entropy} can increase incompatibilities across token values proposed during parallel decoding. Above, \texttt{CPA} should expand to \texttt{Certified Public Accountant}. While unguided generation with parallel decoding method Fast-dLLM \citep{fastdllm} produces the correct token prefix for \texttt{CPA}, Fast-dLLM with guidance fails. Right: Our method is much faster than sequential guided generation while retaining similar generation quality. }
    \label{fig:teaser}
    \vspace{-1.75mm}
\end{figure}

\begin{tcolorbox}[width=\linewidth, sharp corners=all, colback=white!95!black]
{\it Can we accelerate reward-guided dLLM inference while preserving its generation quality?}
\end{tcolorbox}

In this paper, we present, to the best of our knowledge, the first study of accelerating gradient-based reward guidance for dLLMs. Our main contributions are:

\begin{enumerate}[leftmargin=*, topsep=2pt, itemsep=1pt]
    \item We show that, surprisingly, reusing the guidance vector to unmask a group of $k$ tokens is benign. However, reusing the dLLM logits degrades generation quality. Specifically, we observe that (\obsone{}) at masked positions with high confidence scores, the token distribution after guidance remains similar when guidance from an earlier diffusion timestep is reused. Conversely, (\obstwo{}) guidance can increase incompatibilities between co-proposed token values compared to unguided decoding (Figure \ref{fig:teaser}a), preventing reuse of the dLLM logits.
    
    \item We introduce \ours{}, a training-free, hybrid, and adaptive acceleration strategy for reward-guided dLLM decoding. Motivated by \obsone{}, similar to parallel decoding, we select at each iteration a group of candidate token positions with high confidence scores and compute guidance once for the group. To address \obstwo{}, we then autoregressively unmask candidates within this group, recomputing the dLLM logits after each unmasking so that subsequent token choices are conditioned on prior unmaskings. To make these updates efficient, we use sparse recomputation with KV caching \citep{dkvcache}, updating only a small subset of dLLM token representations. Finally, token candidates whose recomputed distributions have low confidence scores are deferred to a later iteration, adaptively balancing the parallel and autoregressive decoding steps.
    
    \item Our guidance reuse principle allows us to extend five existing parallel decoders to the guided setting. Across three reward benchmarks and two dLLMs, \ours{} generates responses up to $4.4\times$ faster than sequential guided decoding while retaining similar reward model scores. 
\end{enumerate}

\section{Background and Related Work}
\label{sec:background}

In this section, we review masked dLLM inference, parallel decoding strategies, and reward guidance methods. Given a prompt $\vp$, the masked dLLM generates a fixed-length response $\vx_0$ of length $L$ where each token comes from a vocabulary $\sV$, where $\sV$ includes a $\mask$ token.

\paragraph{Masked dLLM inference.} A masked dLLM generates $\vx_0$ by
iteratively denoising a partially masked state
$\vx_t\in\sV^L$ \citep{austin2021structured,sahoo2024simple}. We will refer to $t$ as the diffusion timestep, ranging from $T$ to $0$, where $\vx_T$ is a sequence with all masked tokens. At every diffusion timestep $t$, we denote the set of masked and unmasked positions by 
\begin{equation}
\sM_t=\{i\in[L]:x_t^i=\mask\},
\qquad
\sU_t=[L]\setminus\sM_t,
\end{equation}
where $[L]=\{1,\ldots,L\}$. We will use $\vx_t^{\text{u}}$ as a shorthand to represent the set of unmasked tokens at time $t$.
At each denoising step, the dLLM processes the full state $\vx_t$ and
predicts a categorical distribution at every masked position:
\begin{equation}
p_\theta(x_t^i\mid\vx_t^{\text{u}})
=\operatorname{softmax}\!\left(\ell_t^{\text{unguided}}(x_t^i\mid\vx_t^{\text{u}})\right),
\qquad i\in\sM_t.
\end{equation}
Throughout the paper, we will use $\ell_t$ to refer to logits and $p_\theta$ as the corresponding normalized probability distribution. A decoding rule then chooses which token positions to unmask and which token
values to place at those positions. A common confidence-based decoding rule is to pick the token $i$ whose predicted distribution $p_\theta(x_t^i\mid\vx_t^{\text{u}})$ has minimum entropy (which we denote as \textit{high-confidence tokens}), and then sample from that distribution \citep{nie2026large}.

\paragraph{Parallel decoding.} Instead of sampling one token at each diffusion timestep (sequential decoding), parallel decoding methods unmask $k$ tokens at each step, where $k$ can be fixed a priori \citep{nie2026large} or adaptive at inference time \citep{klass,fastdllm}. We refer to these $k$ token positions as $\sA_t=\{a_1,\ldots,a_k\}\subseteq\sM_t$ and $\vc=(c_1,\ldots,c_k)\in\sV^k$ as their proposed values. The true joint distribution over these $k$ tokens would be
\begin{equation} \label{eq:joint}
    p_\theta(x_t^{a_1} = c_1, \dots, x_t^{a_k} = c_k \mid \vx_t^{\text{u}}).
\end{equation}
However, the model only outputs marginal distributions $p_\theta(x_t^{a_j} = c_j \mid \vx_t^{\text{u}})$ at every token position, so a common approximation is to model this joint distribution as a product of marginal distributions $\prod_{j = 1}^k p_\theta(x_t^{a_j} = c_j \mid \vx_t^{\text{u}})$.
%
%
The key challenge is to mitigate this approximation error, known as the joint-marginal mismatch \citep{eb_sampler,liu2025discrete}. For instance, some works address this by selecting positions with limited dependencies using signals such as confidence scores \citep{fastdllm}, analysis of temporal stability of logits \citep{klass,eb_sampler}, or attention-based dependency graphs \citep{kim2026dapd}. However, they are developed for unguided generation, and thus do not address recomputation of the guidance vector as in \Eqref{eq:guidance} or whether the selection rules remain reliable for guided token distributions. An alternative set of approaches uses external planners, such as autoregressive LLMs \citep{israel2026accelerating,liu2025discrete} to autoregressively commit the $k$ tokens. In contrast to these methods, \ours{} is a hybrid decoding strategy that uses parallel decoding techniques for the reward guidance and autoregressive decoding over $\sA_t$. Further, \ours{} obviates the need for external models by efficiently using the dLLM itself to autoregressively unmask tokens. We give a detailed comparison with other decoders in Appendix~\ref{app:related}.

\paragraph{Reward guidance for alignment.} In this work, we focus on the gradient-based reward guidance framework (leaving discussion of other methods to Appendix~\ref{app:related}), which has been shown to outperform methods such as Best-of-N sampling~\citep{liu2025rm,tejaswi2026entropy}. This paradigm draws inspiration from classifier-guidance approaches for continuous diffusion models, which has been remarkably successful in numerous domains for training-free steering \citep{daras2024survey, dhariwal2021diffusion, thaker2025frequency}. For dLLMs, gradient-based guidance methods steer the dLLM logits with a differentiable
sequence-level reward, such as a downstream reward model. As shown in \Eqref{eq:guidance}, the guided logits at each position are the sum of the dLLM logits and a per-token guidance vector $\vr_t$ that shifts the distribution toward higher-reward tokens.
The guidance vector $\vr_t$ is estimated by completing the masked
sequence, evaluating the completion with a reward model, and differentiating
the reward with respect to the dLLM logits. Existing gradient-based guidance methods have focused on improved input completion techniques, e.g., by sampling token values from the predicted marginal distribution at each token \citep{rout2025test}, computing an expected embedding for each position with respect to the predicted distribution at each token \citep{murata2024g2d2}, or interpolating between the two \citep{tejaswi2026entropy}. As such, they have been largely limited to sequential decoding evaluations, resulting in long inference times (Figure \ref{fig:teaser}b). In this work, we focus on the orthogonal problem of accelerating reward guidance since computing forward and backward passes through large reward models at each diffusion timestep can be very costly. 

\section{Method: \ours}
\label{sec:method}

\begin{figure}[t]
    \centering
    \includegraphics[width=\linewidth]{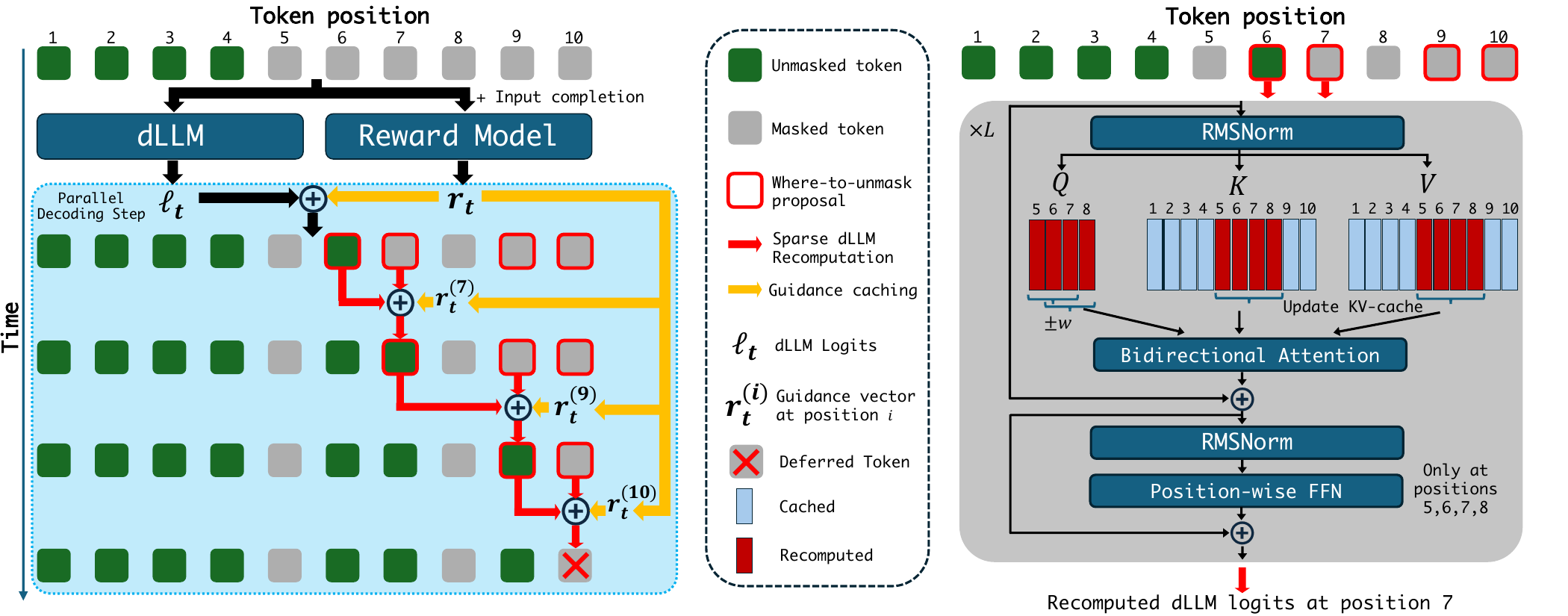}
    \caption{Overview of \ours{}. Left: Illustration of one hybrid decoding step of \ours{} with $k = 4$. \ours{} utilizes guidance caching and autoregressive sparse dLLM recomputation. After recomputation, tokens are deferred if token confidence drops below a threshold. Right: Sparse dLLM recomputation of token 7 after committing token 6, with window radius $w = 1$. Attention and FFN outputs are recomputed only at the selected positions, whose queries attend to the full sequence using updated K/V inside the windows and cached K/V elsewhere.}
    \label{fig:method}
\end{figure}

In this section, we develop \ours{}, an algorithm to accelerate reward-guided decoding for dLLMs. First, in Section~\ref{sec:hybrid}, we derive our hybrid decoding strategy using guidance caching and autoregressive dLLM recomputation. In Section~\ref{sec:sparse_recomp}, we use KV caching and sparse recomputation of attention \citep{dkvcache} to efficiently perform autoregressive dLLM recomputation. Lastly, we introduce confidence deferral in Section~\ref{sec:adaptive}, which is a technique to make decoding adaptive rather than unmasking a fixed number of tokens in each step. Our final method is summarized in Figure \ref{fig:method} and pseudocode is given in Appendix~\ref{app:algorithm}. Throughout this section, we instantiate unguided parallel decoding as the widely used rule of confidence-based unmasking of the top $k$ tokens \citep{nie2026large}, and we instantiate reward guidance as the recently introduced EntRGi algorithm \citep{tejaswi2026entropy}. In Section~\ref{sec:experiments}, we compare \ours{} with other parallel decoders and evaluate other guidance algorithms. Evaluations in this section were performed on RM-Bench \citep{liu2025rm} with the Dream text diffusion model \citep{ye2025dream} and Skywork-Reward-V2-Qwen3-0.6B reward model \citep{liu2026skywork}, with further experimental details given in Appendix \ref{app:method_exp_details}.  

\subsection{Hybrid Decoding: Guidance Caching and dLLM Recomputation}
\label{sec:hybrid}

Recall from \Eqref{eq:guidance} that both the dLLM logits and the guidance vector depend on the unmasked tokens $\vx_t^{\mathrm{u}}$ at each timestep. Existing sequential reward-guided decoders recompute both terms after every unmasking, which requires one dLLM forward pass and one guidance computation per token. Since sequential reward-guided decoding has shown promising performance \citep{tejaswi2026entropy}, we take it as our starting point and seek to maintain its generation quality while reducing its inference time by unmasking $k$ tokens per diffusion timestep. Since \Eqref{eq:guidance} consists of two terms, this raises the question of whether the dLLM logits, the guidance vector, or both can be safely reused to unmask a candidate token set $\sA_t = \{a_1, \dots, a_k\}$ (defined in Section~\ref{sec:background}). 

\begin{wrapfigure}[17]{r}{0.45\linewidth}
    \vspace{-0.5\baselineskip}
    \centering
    \ifrewardfigs\includegraphics[width=\linewidth]{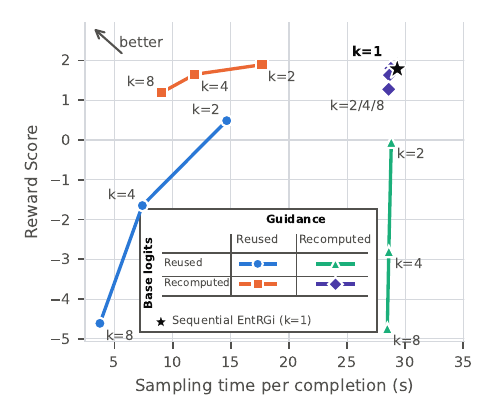}\else\includegraphics[width=\linewidth]{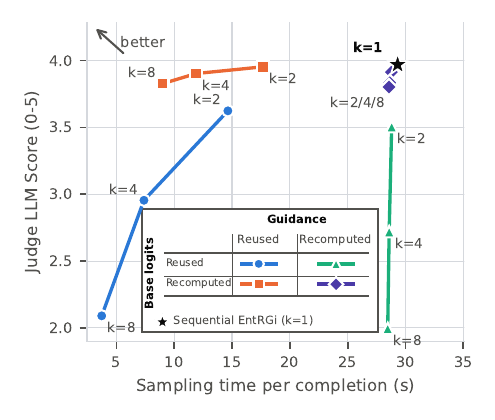}\fi
    \caption{\textbf{Disentangling Reuse of dLLM Logits and Guidance.}}
    \label{fig:refresh-2x2}
    \vspace{-0.5\baselineskip}
\end{wrapfigure}

\paragraph{Separating dLLM and guidance reuse.} To study this question, we independently vary whether each term is reused or recomputed within the candidate token set. Specifically, consider an ordering of $\sA_t$ (e.g., descending confidence scores) and let $\vx_t^{(s)} \in \sV^L$ denote the sequence after unmasking the first $s$ candidates of $\sA_t$, with $\vx_t^{(0)}=\vx_t$. Then, for token position $a_j \in \sA_t$, sequential guided decoding corresponds to iteratively computing for all $j \in \{1,\dots, k\}$:
\begin{equation}
    \operatorname{softmax}\left(\ell_t^{\text{unguided}}\left(x_t^{a_j} \mid (\vx_t^{(s)})^{\mathrm{u}}\right) + \vr_t\left(x_t^{a_j} \mid (\vx_t^{(s')})^{\mathrm{u}}\right)\right)
    \label{eq:parallel-variants}
\end{equation}
for $s = s' = j - 1$. Instead, we analyze variants of \Eqref{eq:parallel-variants} where $s, s' \in \{0,\, j - 1\}$. Here, $s = 0$ (resp. $s' = 0$) denotes reuse of the dLLM logits (resp. guidance vector) computed at the start of the candidate group, whereas $s = j - 1$ (resp. $s' = j - 1$) denotes autoregressive recomputation after each token unmasking. We refer to computing guidance once and reusing it across the candidate group ($s' = 0$) as \emph{guidance caching}, which is particularly attractive because guidance computation dominates inference time (Figure~\ref{fig:teaser}b). When both $s$ and $s'$ are $0$, both terms are sampled from the guided distributions computed at the initial state, following the product-of-marginals approximation of Section~\ref{sec:background}. Figure~\ref{fig:refresh-2x2} demonstrates that guidance caching along with dLLM recomputation retains most of the quality of recomputing both terms (orange vs. purple). Conversely, reusing the dLLM logits (blue and green curves) leads to quality degradation irrespective of guidance caching.

\paragraph{Hybrid decoding.} This analysis motivates combining guidance caching with autoregressive dLLM recomputation, yielding a hybrid decoding strategy. As in parallel decoding, we amortize the guidance computation over $k$ token candidates. Within this candidate set, we unmask candidates autoregressively, recomputing the dLLM logits after each unmasking. Thus, the token at position $a_j$ is chosen from
\begin{equation}
\operatorname{softmax}\left(\ell_t^{\text{unguided}}\left(x_t^{a_j} \mid (\vx_t^{(j-1)})^{\mathrm{u}}\right) + \vr_t\left(x_t^{a_j} \mid (\vx_t^{(0)})^{\mathrm{u}}\right)\right).
\label{eq:our-update}
\end{equation}

\subsubsection{Counterfactual Analysis of Hybrid Decoding}
\label{sec:hybrid-analysis}

\begin{figure}[t]
\centering
\begin{subfigure}[t]{0.45\linewidth}
\centering
\includegraphics[width=\linewidth]{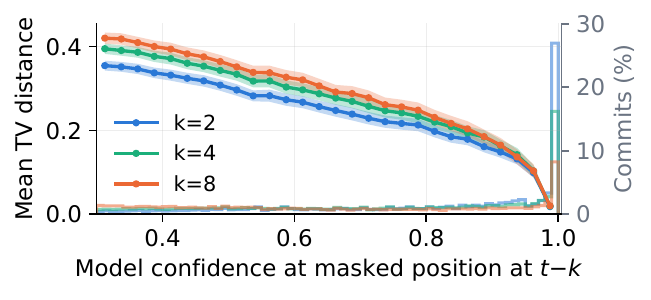}
\phantomcaption
\label{fig:guidance-stability}
\end{subfigure}\hfill
\begin{subfigure}[t]{0.45\linewidth}
\centering
\includegraphics[width=\linewidth]{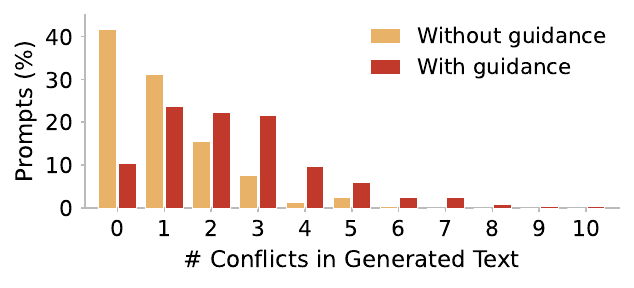}
\phantomcaption
\label{fig:guidance-incompatibility}
\end{subfigure}
\vspace{-0.5\baselineskip}
\caption{Left: (a) Replacing fresh guidance with a cached guidance produces small changes in high-confidence token distributions. Right: (b) Percentage of prompts exhibiting certain number of conflicts between proposed tokens (as measured by PMI being less than $-1$). }
\label{fig:guidance-obs} 
\end{figure}

To better understand the asymmetry between guidance reuse and dLLM logits reuse, we run two counterfactual studies along sequential decoding trajectories (Figure~\ref{fig:guidance-obs}).

\textbf{Observation 1 (\obsone{}): guidance reuse has little effect at high-confidence positions.} At every diffusion timestep, we hold the current dLLM logits fixed and replace the guidance vector with one cached from an earlier step. We measure the total-variation distance between the resulting guided distributions (Figure~\ref{fig:guidance-stability}). We find that this distance is smaller at high-confidence positions, which are precisely the positions that standard decoding strategies unmask \citep{ye2025dream}. This helps explain why guidance caching is effective.

\textbf{Observation 2 (\obstwo{}): guidance can make co-proposed tokens locally incompatible.} Tokens proposed in parallel from the same partially masked sequence need not be likely under the true joint distribution of the tokens since the model only outputs marginal distributions for each token. While this joint-marginal mismatch is present even without guidance \citep{eb_sampler}, we find that guidance can exacerbate this mismatch. To measure the effect of guidance on the joint-marginal mismatch, we consider sequential unguided trajectories and toggle guidance at each state, examining its effect on the top-2 confident tokens that would be committed in parallel. We quantify the joint-marginal mismatch at these two tokens as the dLLM-predicted Pointwise Mutual Information (PMI), which is the difference in log probabilities of the joint $p(a, b) = p(a) p(b \mid a)$ and product of marginal distributions $p(a)p(b)$. Negative PMI indicates that committing the first token lowers the dLLM-predicted probability of the second relative to its marginal probability. Figure~\ref{fig:guidance-incompatibility} shows that guidance actually increases incompatibility among proposed tokens. The average number of incompatible pairs rises from $1.04$ without guidance to $2.47$ with guidance, and the fraction of prompts with at least one incompatible pair rises from $58\%$ to $90\%$. This illustrates the role of dLLM recomputation in our hybrid decoding strategy as incompatible proposals can be revised before unmasking.

\subsubsection{Sparse dLLM Recomputation}
\label{sec:sparse_recomp}

Despite the speedups obtained from reducing guidance computations, exact dLLM recomputation still requires the same number of dLLM forward passes as sequential decoding. To reduce this cost, we utilize dLLM acceleration techniques \citep{dkvcache,dllmcache}. Specifically, for every hybrid decoding step, once the $k$ candidate positions are fixed, recomputing logits for the next candidate requires updated logits only at its position $a_j$. Thus, we make a connection to \citet{dkvcache}, where the authors fix the token generation order of unguided sequential dLLM decoding to develop efficient sparse dLLM forward passes. Specifically, we can approximate the full forward pass by recomputing attention and feed-forward outputs for every transformer layer at a small set of positions: the union of windows of radius $w$ around the most recently committed token and the next candidate position. To recompute attention, the queries at only these positions attend to the full sequence, using newly computed keys and values inside the windows and cached keys and values elsewhere (see Figure~\ref{fig:method}). This reduces the time complexity of each recomputation from $O(L^2)$ to $O(wL)$. The updated keys and values overwrite the cache for subsequent recomputation passes. The sparse dLLM forward produces approximate base logits $\widehat\ell_t^{\text{unguided}}\left(x_t^{a_j} \mid (\vx_t^{(j-1)})^{\mathrm{u}}\right)$, which we combine with the cached reward gradient as in \Eqref{eq:our-update}. 

\subsection{Adaptivity via Confidence Deferral}
\label{sec:adaptive}

Unmasking exactly $k$ tokens in every step can be restrictive \citep{fastdllm} since the number of tokens that are safe to unmask together depends on the prompt and decoding state. We therefore introduce an adaptive confidence deferral strategy. Specifically, after sparse dLLM recomputation, we compute
\begin{equation}
    \gamma_j=\max_{v \in \sV} \operatorname{softmax}\left(\widehat\ell_t^{\text{unguided}}\left(x_t^{a_j} \mid (\vx_t^{(j-1)})^{\mathrm{u}}\right) + \vr_t\left(x_t^{a_j} \mid (\vx_t^{(0)})^{\mathrm{u}}\right)\right)(v).
    \label{eq:verified-confidence}
\end{equation}
If $\gamma_j\ge\tau$ for a chosen threshold $\tau$, we commit the verified token. If not, the position remains masked and is reconsidered in a later step, when both the dLLM logits and the guidance are recomputed. The first token candidate is always committed, so every step makes progress. This makes the degree of parallelism adaptive and postpones low-confidence candidates until a later step so both dLLM logits and guidance can be recomputed.

In summary, for each decoding step, \ours{} performs $1$ dLLM forward pass, $1$ guidance computation step, and $k - 1$ sparse recomputation passes, as opposed to the $k$ dLLM forward passes and $k$ guidance computations for sequential guided decoding. Refer to Appendix~\ref{app:algorithm} for psuedocode.

\section{Experiments}
\label{sec:experiments}

Below, we describe our experimental setup, deferring other experimental details to Appendix \ref{app:details}.
\begin{table}[t]
\begin{center}
\small
\setlength{\tabcolsep}{2.5pt}
\begin{tabular}{c@{\hspace{2pt}}lcccc c cccc}
& & \multicolumn{4}{c}{\bf Dream-7B} & & \multicolumn{4}{c}{\bf LLaDA-8B} \\
\cmidrule(lr){3-6} \cmidrule(lr){8-11}
\multicolumn{2}{c}{\bf Method} & \bf Top@1 & \bf Seq Gap & \bf LMUnit & \bf s/gen & & \bf Top@1 & \bf Seq Gap & \bf LMUnit & \bf s/gen \\
\hline \hline
\multicolumn{11}{c}{\emph{JudgeBench}} \\
\hline \hline
\g  & Conf (BoN) & $+1.52_{\pm .16}$ & -- & $3.97_{\pm .03}$ & $10.8_{\pm 4.7}$ & & $+2.65_{\pm .21}$ & -- & $4.05_{\pm .03}$ & $11.8_{\pm 5.1}$ \\
\g \multirow{-2}{*}{\rotatebox[origin=c]{90}{\footnotesize\scshape Seq}} & Conf & $+2.53_{\pm .16}$ & -- & $4.05_{\pm .03}$ & $37.3_{\pm 15.5}$ & & $+3.23_{\pm .20}$ & -- & $4.08_{\pm .03}$ & $45.3_{\pm 18.5}$ \\
\hline
\multirow{7}{*}{\rotatebox[origin=c]{90}{\footnotesize\scshape Parallel}} & Conf (BoN) & $-3.14_{\pm .20}$ & $-4.66$ & $3.01_{\pm .05}$ & $\phantom{0}2.7_{\pm 1.1}$ & & $-2.23_{\pm .22}$ & $-4.88$ & $3.05_{\pm .05}$ & $\phantom{0}3.0_{\pm 1.3}$ \\
 & Conf & $+0.38_{\pm .20}$ & $-2.15$ & $3.64_{\pm .04}$ & $\phantom{0}9.7_{\pm 3.8}$ & & $+1.54_{\pm .22}$ & $-1.69$ & $3.78_{\pm .04}$ & $11.8_{\pm 5.1}$ \\
 & Fast-dLLM & $+0.26_{\pm .21}$ & $-2.27$ & $3.62_{\pm .05}$ & $\phantom{0}7.9_{\pm 4.1}$ & & $-0.12_{\pm .21}$ & $-3.35$ & $3.39_{\pm .05}$ & $\phantom{0}6.5_{\pm 2.9}$ \\
 & KLASS & $-0.06_{\pm .22}$ & $-2.59$ & $3.61_{\pm .05}$ & $10.8_{\pm 12.5}$ & & $-0.81_{\pm .19}$ & $-4.04$ & $3.10_{\pm .05}$ & $\phantom{0}5.6_{\pm 2.7}$ \\
 & EB-Sampler & $-0.49_{\pm .19}$ & $-3.02$ & $3.49_{\pm .04}$ & $\phantom{0}9.4_{\pm 3.5}$ & & $+0.09_{\pm .21}$ & $-3.14$ & $3.52_{\pm .04}$ & $10.1_{\pm 3.8}$ \\
 & DAPD & $+1.92_{\pm .17}$ & $-0.61$ & $3.95_{\pm .03}$ & $\phantom{0}8.6_{\pm 3.3}$ & & $+2.74_{\pm .22}$ & $-0.49$ & $3.99_{\pm .04}$ & $\phantom{0}9.4_{\pm 4.2}$ \\
 & \ours{} & $\mathbf{+2.08}_{\pm .17}$ & $\mathbf{-0.45}$ & $\mathbf{4.02}_{\pm .03}$ & $\phantom{0}9.6_{\pm 5.0}$ & & $\mathbf{+2.76}_{\pm .21}$ & $\mathbf{-0.47}$ & $\mathbf{4.03}_{\pm .03}$ & $10.3_{\pm 3.4}$ \\
\hline \hline
\multicolumn{11}{c}{\emph{RM-Bench}} \\
\hline \hline
\g  & Conf (BoN) & $+4.41_{\pm .12}$ & -- & $4.06_{\pm .02}$ & $\phantom{0}6.3_{\pm 1.9}$ & & $+4.39_{\pm .12}$ & -- & $3.94_{\pm .02}$ & $\phantom{0}6.6_{\pm 2.2}$ \\
\g \multirow{-2}{*}{\rotatebox[origin=c]{90}{\footnotesize\scshape Seq}} & Conf & $+5.42_{\pm .11}$ & -- & $4.15_{\pm .02}$ & $22.9_{\pm 6.9}$ & & $+5.25_{\pm .11}$ & -- & $4.02_{\pm .02}$ & $24.9_{\pm 8.8}$ \\
\hline
\multirow{7}{*}{\rotatebox[origin=c]{90}{\footnotesize\scshape Parallel}} & Conf (BoN) & $-1.96_{\pm .14}$ & $-6.37$ & $2.75_{\pm .02}$ & $\phantom{0}1.6_{\pm .5}$ & & $-1.57_{\pm .14}$ & $-5.96$ & $2.79_{\pm .02}$ & $\phantom{0}1.7_{\pm .5}$ \\
 & Conf & $+2.32_{\pm .14}$ & $-3.09$ & $3.61_{\pm .02}$ & $\phantom{0}5.8_{\pm 1.7}$ & & $+2.83_{\pm .13}$ & $-2.43$ & $3.63_{\pm .02}$ & $\phantom{0}6.3_{\pm 2.2}$ \\
 & Fast-dLLM & $+1.34_{\pm .15}$ & $-4.08$ & $3.51_{\pm .03}$ & $\phantom{0}5.6_{\pm 2.2}$ & & $+0.74_{\pm .14}$ & $-4.51$ & $3.31_{\pm .03}$ & $\phantom{0}4.5_{\pm 1.7}$ \\
 & KLASS & $+0.38_{\pm .15}$ & $-5.04$ & $3.33_{\pm .03}$ & $\phantom{0}9.6_{\pm 9.9}$ & & $-0.11_{\pm .14}$ & $-5.36$ & $3.03_{\pm .03}$ & $\phantom{0}4.1_{\pm 1.9}$ \\
 & EB-Sampler & $+0.40_{\pm .15}$ & $-5.01$ & $3.20_{\pm .02}$ & $\phantom{0}5.9_{\pm 2.0}$ & & $+0.49_{\pm .15}$ & $-4.76$ & $3.19_{\pm .02}$ & $\phantom{0}5.9_{\pm 1.6}$ \\
 & DAPD & $+3.58_{\pm .13}$ & $-1.83$ & $3.87_{\pm .02}$ & $\phantom{0}5.9_{\pm 1.3}$ & & $+4.36_{\pm .12}$ & $-0.89$ & $3.90_{\pm .02}$ & $\phantom{0}6.2_{\pm 2.3}$ \\
 & \ours{} & $\mathbf{+4.70}_{\pm .12}$ & $\mathbf{-0.72}$ & $\mathbf{4.04}_{\pm .02}$ & $\phantom{0}6.2_{\pm 1.5}$ & & $\mathbf{+4.53}_{\pm .12}$ & $\mathbf{-0.72}$ & $\mathbf{3.92}_{\pm .02}$ & $\phantom{0}6.5_{\pm 1.8}$ \\
\hline \hline
\multicolumn{11}{c}{\emph{Reward-Bench-2}} \\
\hline \hline
\g  & Conf (BoN) & $+2.60_{\pm .10}$ & -- & $4.18_{\pm .01}$ & $\phantom{0}5.5_{\pm 1.4}$ & & $+2.19_{\pm .09}$ & -- & $4.00_{\pm .02}$ & $\phantom{0}6.1_{\pm 1.6}$ \\
\g \multirow{-2}{*}{\rotatebox[origin=c]{90}{\footnotesize\scshape Seq}} & Conf & $+3.39_{\pm .10}$ & -- & $4.23_{\pm .01}$ & $20.4_{\pm 5.2}$ & & $+2.88_{\pm .09}$ & -- & $4.05_{\pm .02}$ & $26.9_{\pm 9.3}$ \\
\hline
\multirow{7}{*}{\rotatebox[origin=c]{90}{\footnotesize\scshape Parallel}} & Conf (BoN) & $-1.98_{\pm .09}$ & $-4.58$ & $2.95_{\pm .02}$ & $\phantom{0}1.4_{\pm .4}$ & & $-2.22_{\pm .08}$ & $-4.41$ & $2.88_{\pm .02}$ & $\phantom{0}1.5_{\pm .4}$ \\
 & Conf & $+1.24_{\pm .10}$ & $-2.15$ & $3.78_{\pm .02}$ & $\phantom{0}5.2_{\pm 1.2}$ & & $+0.98_{\pm .09}$ & $-1.90$ & $3.67_{\pm .02}$ & $\phantom{0}6.6_{\pm 2.3}$ \\
 & Fast-dLLM & $+1.36_{\pm .11}$ & $-2.03$ & $3.82_{\pm .02}$ & $\phantom{0}6.8_{\pm 2.5}$ & & $+0.05_{\pm .09}$ & $-2.83$ & $3.55_{\pm .02}$ & $\phantom{0}7.9_{\pm 4.0}$ \\
 & KLASS & $+1.06_{\pm .10}$ & $-2.33$ & $3.83_{\pm .02}$ & $15.7_{\pm 13.3}$ & & $-0.09_{\pm .09}$ & $-2.97$ & $3.44_{\pm .02}$ & $\phantom{0}8.0_{\pm 4.4}$ \\
 & EB-Sampler & $+0.56_{\pm .10}$ & $-2.83$ & $3.57_{\pm .02}$ & $\phantom{0}6.6_{\pm 2.0}$ & & $+0.01_{\pm .09}$ & $-2.87$ & $3.40_{\pm .02}$ & $\phantom{0}9.0_{\pm 3.1}$ \\
 & DAPD & $+3.01_{\pm .10}$ & $-0.38$ & $4.15_{\pm .01}$ & $\phantom{0}6.2_{\pm 1.7}$ & & $+2.49_{\pm .09}$ & $-0.39$ & $\mathbf{4.00}_{\pm .02}$ & $\phantom{0}7.2_{\pm 2.2}$ \\
 & \ours{} & $\mathbf{+3.25}_{\pm .10}$ & $\mathbf{-0.14}$ & $\mathbf{4.19}_{\pm .01}$ & $\phantom{0}7.3_{\pm 2.1}$ & & $\mathbf{+2.58}_{\pm .09}$ & $\mathbf{-0.30}$ & $\mathbf{4.00}_{\pm .02}$ & $\phantom{0}7.3_{\pm 2.5}$ \\
\end{tabular}
\end{center}
\caption{Quantitative results for reward-guided generation on JudgeBench,
RM-Bench, and Reward-Bench-2. All
parallel decoding methods use cached EntRGi reward guidance, unless (BoN) is mentioned, denoting Best-of-N sampling. Subscripts denote the standard error over prompts.}
\label{tab:guided4}
\end{table}

\paragraph{Benchmarks.} Following prior work on reward guidance, we evaluate \ours{} on three reward benchmarks (JudgeBench \citep{tan2025judgebench}, RM-Bench \citep{liu2025rm}, and Reward-Bench-2 \citep{malik2025rewardbench2}), spanning various categories such as Coding, Math, Precise Instruction Following (IF), and Safety. 

\paragraph{Baselines.} We evaluate \ours{} on two popular open-source dLLMs: Dream-7B-Instruct \citep{ye2025dream} and LLaDA-8B-Instruct \citep{nie2026large}. While existing parallel decoding baselines are evaluated on unguided decoding, our guidance caching strategy (Section \ref{sec:method}) allows us to extend existing methods to the guided setting. To our knowledge, our work represents the first evaluation of these decoders in the reward-guided setting. We compare against several training-free parallel decoding baselines such as confidence-based decoding (which we abbreviate as Conf.) \citep{nie2026large}, Fast-dLLM \citep{fastdllm}, KLASS \citep{klass}, EB-sampler \citep{eb_sampler} and DAPD \citep{kim2026dapd}. For fair comparison, we do not utilize additional optimizations of the full dLLM forward pass such as prompt caching \citep{fastdllm}, which are orthogonal improvements and can be combined with \ours{}. We also compare to Best-Of-N sampling with $N=4$. We use the recent state-of-the-art reward guidance method EntRGI \citep{tejaswi2026entropy} with the Skywork-Reward-V2-Qwen3-1.7B reward model \citep{liu2026skywork}. We defer evaluation of other guidance methods and different reward models to Appendix \ref{app:guidance-general} and \ref{app:rm-ablation}.

\paragraph{Metrics.} Following \citet{tejaswi2026entropy}, we generate $4$ trajectories for each prompt and measure generation quality by the reward model score. We report Top@1 (the maximum reward model score across all trajectories) and Seq Gap (the gap between parallel decoding Top@1 and the corresponding sequential decoding baseline using the same guidance algorithm). To monitor reward overoptimization \citep{gao2023scaling}, we report LMUnit (external judge LLM LMUnit-Qwen2.5-72B \citep{saad2025lmunit} score from $1-5$). We further report Avg@4 (the average reward model score across the $4$ trajectories) in Appendix \ref{app:full-tables}. To evaluate inference time, we report s/gen (wall-clock inference time for each generation), computed on a Nvidia A5000 GPU. 

\paragraph{Hyperparameters.}  We set the generation length to be $128$ for all methods, and baseline hyperparameters can be found in Appendix \ref{app:hparams}. \ours{} uses a candidate pool of $k=8$ for all experiments along with confidence deferral parameter $\tau = 0.5$ and attention window size $w=2$. We consider a higher throughput regime by increasing $k$ to $16$ in Appendix~\ref{app:additional}. We use batch size 4 for Dream-7B and batch size 2 for LLaDA-8B.

\subsection{Main Results}

\begin{figure}[t]
\begin{minipage}[b]{0.5\linewidth}
\centering
\includegraphics[width=\linewidth]{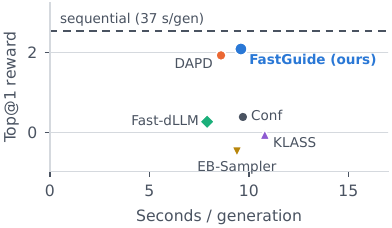}
\caption{Quality versus inference time on JudgeBench using Dream-7B model.}
\label{fig:frontier}
\end{minipage}\hfill
\begin{minipage}[b]{0.46\linewidth}
\centering
\ifrewardfigs\includegraphics[width=\linewidth]{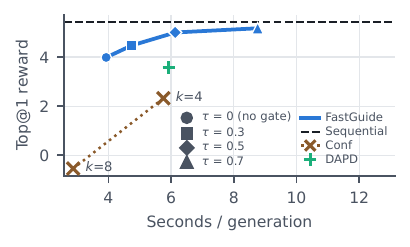}\else\includegraphics[width=\linewidth]{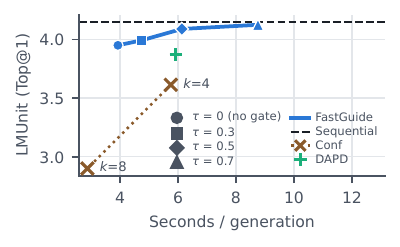}\fi
\caption{$\tau$ sweep for confidence deferral on RM-Bench using Dream-7B model.}
\label{fig:tau-sweep-compact}
\end{minipage}
\vspace{-0.4\baselineskip}
\end{figure}

\paragraph{Reward vs Inference Time.} Table \ref{tab:guided4} and Figure \ref{fig:frontier} show our main results. \ours{} is $2.8$--$4.4\times$ faster than sequential guided decoding across the three benchmarks and two models while remaining within 0.75 reward points of sequential guidance on both models. Its high LMUnit scores provide complementary evidence of response quality and also guard against reward overoptimization. Compared to unguided sequential Best-of-N sampling, sequential guidance improves Top@1 by $0.6$--$1.0$ reward points but costs $3.5$--$4.4\times$ the inference time. \ours{} reduces this cost to $0.9$--$1.3\times$ that of Best-of-N while retaining a Top@1 improvement of $0.1$--$0.65$ reward points. These conclusions also hold on LLaDA-8B, where \ours{} again attains the smallest Seq Gap, while Fast-dLLM and KLASS degrade even more severely than on Dream-7B. Cached guidance on top of confidence-based decoding (Conf row) improves Top@1 by roughly $4$ reward points over its unguided counterpart (Conf (BoN) row), confirming that guidance caching preserves the benefit of guidance. FastGuide further narrows this gap through sparse dLLM recomputation and confidence deferral. An important benefit of \ours{} is that sparse dLLM recomputation transforms the compute-bound full dLLM forward pass, that involves 
bidirectional attention over the whole sequence, into a memory-bound sparse forward over a few positions. Thus, standard techniques such as batching can reduce inference time per trajectory, which we explore in Appendix \ref{app:batching}.

\par\ifdim\dimexpr\pagegoal-\pagetotal\relax<14\baselineskip\newpage\fi
\begin{wrapfigure}[14]{r}{0.57\linewidth}
\vspace{-\intextsep}
\captionsetup{skip=3pt}
\centering
\includegraphics[width=\linewidth]{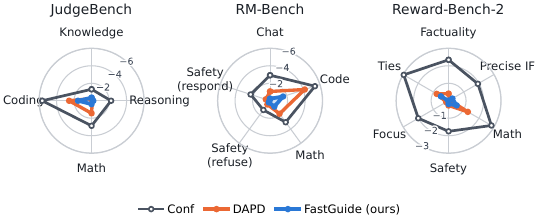}
\caption{\textbf{Sequential gap by category (closer to $0$ is better).} \ours{} has the smallest gap to sequential decoding, especially in reasoning categories such as coding where baselines degrade.}
\label{fig:seqgap-radar}
\vspace{-0.5\baselineskip}
\end{wrapfigure}
\paragraph{Comparison to Parallel Decoders.} Most existing parallel decoders lose a large part of the benefit of guidance, with Fast-dLLM, KLASS, and EB-sampler suffering substantial performance degradation relative to sequential decoding. In contrast, DAPD, which selects positions from the dLLM's attention graph rather than from marginal confidence, is by far the strongest baseline. \ours{} matches or improves on DAPD on every benchmark and model, with the largest gains on RM-Bench ($+1.1$ Top@1 on Dream-7B). Figure~\ref{fig:seqgap-radar} analyzes this Seq Gap further by splitting performance along benchmark categories. DAPD exhibits larger reward gaps on coding and math tasks. This pattern is consistent with the greater sensitivity to incompatible token choices in these categories, and the qualitative example in Figure~\ref{fig:qual-boxes} further illustrates this. In contrast, for categories such as chat and safety, an incompatible pair of tokens costs little fluency.  \ours{} has a small Seq Gap across nearly all categories, highlighting the benefit of sparse dLLM recomputation in the reward-guided setting. 

\paragraph{Qualitative Samples.} Figure~\ref{fig:qual-boxes} shows an illustrative mathematical reasoning prompt from RM-Bench. Unguided DAPD has a correct reasoning trace, but under guidance it commits an incorrect line of arithmetic reasoning ($5^2$ in place of $8^2$) that the rest of the derivation then propagates to a wrong final answer, showing an instance of \obstwo{}. Instead, \ours{} recomputes each candidate token distribution before unmasking it and recovers the correct answer. The reward model scores in Figure~\ref{fig:qual-boxes} reflect this ordering. Appendix~\ref{app:qual} shows further qualitative examples.
\definecolor{qbase}{HTML}{D6453D}
\definecolor{qours}{HTML}{3B7DD8}
\definecolor{qerr}{HTML}{B3261E}
\definecolor{qerrbg}{HTML}{FBE3E1}
\newcommand{\qerrtok}[1]{{\setlength{\fboxsep}{0.5pt}\colorbox{qerrbg}{\textcolor{qerr}{#1}}}}
\newtcolorbox{qualbox}[2]{enhanced, colback=white, colframe=#1, boxrule=0.6pt,
  arc=2pt, left=4pt, right=4pt, top=7pt, bottom=2.5pt, boxsep=0pt,
  before skip=2pt, after skip=4pt, fontupper=\ttfamily\tiny, parbox=false,
  attach boxed title to top left={xshift=4pt, yshift=-\tcboxedtitleheight/2},
  boxed title style={colback=white, colframe=#1, boxrule=0.6pt, arc=2pt, left=2.5pt, right=2.5pt,
    top=0.8pt, bottom=0.8pt, boxsep=0pt},
  title={\sffamily\scriptsize\textcolor{#1}{#2}}}
\begin{figure}[t]
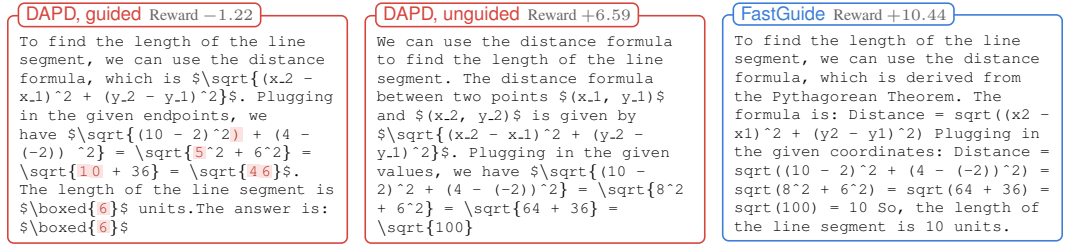

\centering
\begin{minipage}{\linewidth}
{\sffamily\scriptsize\textbf{Prompt:} If the endpoints of a line segment are (2, -2) and (10, 4), what is the length of the segment?}
\par\vspace{2pt}
\begin{minipage}[t]{0.323\linewidth}\vspace{0pt}
\begin{qualbox}{qbase}{DAPD, guided\hspace{3pt}{\tiny\textcolor{black!60}{\textrm{Reward $-1.22$}}}}
To\ find\ the\ length\ of\ the\ line\ segment,\ we\ can\ use\ the\ distance\ formula,\ which\ is\ \$\textbackslash{}sqrt\{(x\_2\ -\ x\_1)\^{}2\ +\ (y\_2\ -\ y\_1)\^{}2\}\$.\ Plugging\ in\ the\ given\ endpoints,\ we\ have\ \$\textbackslash{}sqrt\{(10\ -\ 2)\^{}2\qerrtok{)}\ +\ (4\ -\ (-2))\ \^{}2\}\ =\ \textbackslash{}sqrt\{\qerrtok{5}\^{}2\ +\ 6\^{}2\}\ =\ \textbackslash{}sqrt\{\qerrtok{1}\qerrtok{0}\ +\ 36\}\ =\ \textbackslash{}sqrt\{\qerrtok{4}\qerrtok{6}\}\$.\ The\ length\ of\ the\ line\ segment\ is\ \$\textbackslash{}boxed\{\qerrtok{6}\}\$\ units.The\ answer\ is:\ \$\textbackslash{}boxed\{\qerrtok{6}\}\$
\end{qualbox}
\end{minipage}\hfill
\begin{minipage}[t]{0.323\linewidth}\vspace{0pt}
\begin{qualbox}{qbase}{DAPD, unguided\hspace{3pt}{\tiny\textcolor{black!60}{\textrm{Reward $+6.59$}}}}
We\ can\ use\ the\ distance\ formula\ to\ find\ the\ length\ of\ the\ line\ segment.\ The\ distance\ formula\ between\ two\ points\ \$(x\_1,\ y\_1)\$\ and\ \$(x\_2,\ y\_2)\$\ is\ given\ by\ \$\textbackslash{}sqrt\{(x\_2\ -\ x\_1)\^{}2\ +\ (y\_2\ -\ y\_1)\^{}2\}\$.\ Plugging\ in\ the\ given\ values,\ we\ have\ \$\textbackslash{}sqrt\{(10\ -\ 2)\^{}2\ +\ (4\ -\ (-2))\^{}2\}\ =\ \textbackslash{}sqrt\{8\^{}2\ +\ 6\^{}2\}\ =\ \textbackslash{}sqrt\{64\ +\ 36\}\ =\ \textbackslash{}sqrt\{100\}
\end{qualbox}
\end{minipage}\hfill
\begin{minipage}[t]{0.323\linewidth}\vspace{0pt}
\begin{qualbox}{qours}{FastGuide\hspace{3pt}{\tiny\textcolor{black!60}{\textrm{Reward $+10.44$}}}}
To\ find\ the\ length\ of\ the\ line\ segment,\ we\ can\ use\ the\ distance\ formula,\ which\ is\ derived\ from\ the\ Pythagorean\ Theorem.\ The\ formula\ is:\ Distance\ =\ sqrt((x2\ -\ x1)\^{}2\ +\ (y2\ -\ y1)\^{}2)\ Plugging\ in\ the\ given\ coordinates:\ Distance\ =\ sqrt((10\ -\ 2)\^{}2\ +\ (4\ -\ (-2))\^{}2)\ =\ sqrt(8\^{}2\ +\ 6\^{}2)\ =\ sqrt(64\ +\ 36)\ =\ sqrt(100)\ =\ 10\ So,\ the\ length\ of\ the\ line\ segment\ is\ 10\ units.\ 
\end{qualbox}
\end{minipage}
\end{minipage}
\caption{\textbf{Qualitative generations on Math RM-Bench prompt.} Red highlighted tokens denote incorrect tokens. This reinforces that guidance can worsen performance of parallel decoders, while \ours{} succeeds.}
\label{fig:qual-boxes}
\end{figure}

\subsection{Ablation Studies}
\label{sec:ablations}

\paragraph{Effect of confidence deferral.} Figure~\ref{fig:tau-sweep-compact} sweeps the deferral threshold $\tau$ of \ours{} on RM-Bench, showing that $\tau$ provides an extra knob for interpolating the quality-efficiency tradeoff. Specifically, without deferral ($\tau{=}0$), unmasking a fixed $8$ tokens per step is fastest, but generation quality degrades. As more tokens are deferred, quality increases, approaching the performance of sequential decoding, but at the expense of inference time due to increased guidance computation steps and full dLLM forwards.

\begin{figure}[t]
\centering
\captionsetup{skip=3pt}
\begin{subfigure}[t]{0.485\linewidth}
\centering
\includegraphics[width=\linewidth,trim=0 10 0 0,clip]{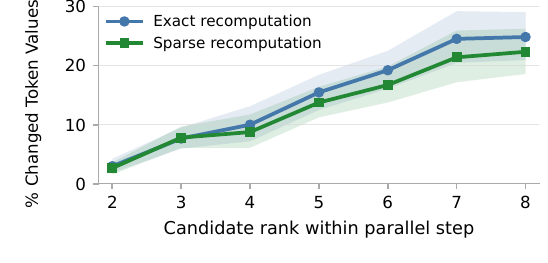}
\caption{}
\label{fig:verification-revisions}
\end{subfigure}\hfill
\begin{subfigure}[t]{0.485\linewidth}
\centering
\includegraphics[width=\linewidth,trim=0 10 0 0,clip]{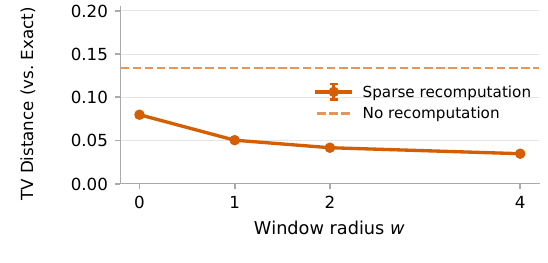}
\caption{}
\label{fig:verification-fidelity}
\end{subfigure}
\caption{\textbf{Analysis of sparse dLLM recomputation vs. exact dLLM recomputation.}}
\label{fig:verification-diagnostics}
\vspace{-0.4\baselineskip}
\end{figure}

\paragraph{Effect of dLLM Recomputation.}
Figure~\ref{fig:verification-diagnostics} analyzes the sparse dLLM recomputation step of \ours{}. Specifically, Figure~\ref{fig:verification-diagnostics}b shows that sparse dLLM recomputation is a useful approximation of the full dLLM forward pass as measured by the TV distance between sparsely and exactly recomputed token distributions, while Figure~\ref{fig:verification-diagnostics}a shows that both sparse and exact recomputation change a large fraction of token values during decoding.  Finally,
Appendix~\ref{app:verify-order} shows that generation quality is largely
agnostic to the order of the candidate set for which autoregressive decoding is performed.

\section{Conclusion}
\label{sec:conclusion}

In this paper, we introduced \ours{}, a training-free hybrid decoding algorithm to accelerate reward-guidance for diffusion language models. Our findings show that while guidance can be reused across diffusion timesteps, guidance can worsen the token incompatibilities of parallel decoding. To address this, \ours{} performs guidance caching combined with sparse dLLM recomputation to autoregressively and efficiently recompute each proposed token distribution in a decoding step. \ours{} achieves speedups up to $4.4\times$ over sequential guided decoding while retaining similar generation quality. To the best of our knowledge, our work is the first comprehensive study of accelerating training-free, gradient-based reward guidance for diffusion language models and the first evaluation of parallel decoders under reward guidance. We hope the ideas proposed in this paper will inspire future work on developing parallel decoders specifically for the guided setting.






\bibliography{iclr2027_conference}
\bibliographystyle{iclr2027_conference}

\clearpage
\appendix
\renewcommand{\contentsname}{Appendix Contents}
\addtocontents{toc}{\protect\setcounter{tocdepth}{2}}
\tableofcontents
\section{Extended Related Work}
\label{app:related}

\paragraph{Other reward guidance methods.} Aligning dLLM outputs with a reward can be done by fine-tuning or at inference time. Training-based approaches adapt RLHF-style objectives \citep{ouyang2022training,rafailov2023direct} to dLLMs, e.g., via policy-gradient or preference optimization \citep{zhao2025d1,zhu2025llada15}, but require retraining whenever the reward changes. Training-free approaches instead steer a frozen dLLM at inference time and split into two families. Sampling-based methods such as Best-Of-N or Sequential Monte Carlo (SMC) methods only require reward evaluations, but can require many sequences/particles to be decoded \citep{wu2023practical,zhao2024probabilistic,singhal2025general,skreta2025feynman,pani2025test}. Gradient-based methods instead differentiate the reward model with respect to the dLLM logits and steer a single trajectory, at the cost of a forward and backward pass through the reward model at every diffusion timestep \citep{murata2024g2d2,rout2025test,tejaswi2026entropy}. One of the main challenges of reward guidance is obtaining gradients since the tokens in the input sequence are discrete. To tackle this, existing methods propose to use stop-gradient approximations \citep{rout2025test} or take gradients with respect to the continuous token embeddings \citep{murata2024g2d2, tejaswi2026entropy}. 

\paragraph{Accelerating guidance.} In continuous diffusion, the cost of guidance has been reduced by only computing classifier gradients on a subset of timesteps \citep{dinh2024compress}, by skipping the classifier-free guidance residual \citep{castillo2023adaptive,lv2024fastercache,yan2025lazymar}, or by replacing reward backpropagation with a closed-form estimate \citep{kim2026lidar}. Our guidance caching has two key differences. In continuous diffusion, the state evolves smoothly with the noise level, so guidance reuse is controlled by a timestep schedule. In contrast, in masked dLLMs, the effect of guidance reuse is position-dependent as well as timestep-dependent. While guidance may change across the entire sequence, \obsone{} demonstrates that at high-confidence positions (which are precisely the positions that are unmasked), the guided token distribution is stable. Second, our work provides a novel observation as to how guidance interacts with the joint-marginal mismatch (\obstwo{}). For traditional autoregressive LLMs, segment-level and blockwise methods amortize reward-model evaluations over several tokens \citep{li2024cards,mudgal2023controlled,liao2025rsd}, but they use the reward as a score for search or rejection rather than as a gradient, and further do not face the joint-marginal mismatch since tokens are generated sequentially.

\paragraph{Parallel decoding.} We now review other parallel decoders. The first class uses external models to supply joint structure to mitigate the joint-marginal mismatch. For instance, \citet{israel2026accelerating}, \citet{liu2025discrete}, and \citet{xu2024edlm} use the structure from an autoregressive verifier, a copula, or an energy function respectively. However, in the reward-guided setting, this requires the auxiliary model to be consistent with the guided distribution, or yield similarly high-reward results. The second class is training-free self-verification methods. Self-speculative decoders \citep{gao2025ssd,wu2025freedave,agrawal2025spiffy,han2026s2d2,cui2026simsd} perform parallel generation by drafting several tokens and verifying each draft conditioned on the earlier ones, using batched full forward passes or a single pass over a duplicated sequence. However, the goal of these methods is to remain lossless by using rejection sampling techniques, which also do not cleanly extend to the reward-guided setting. The final class of methods is revocable parallel decoders \citep{wang2025remdm,hong2025wino,xiang2026cover}, which unmask first and re-mask tokens that fail a verification step. We emphasize that these methods have not been evaluated under gradient-based reward guidance, which presents its own challenges of guidance recomputation and the influence of guidance on token compatibilities (Section \ref{sec:method}).

\ours{} differs from these methods in two main ways. First, \ours{} trades exactness for cost: rather than relying on an auxiliary model or on full forward passes to remain lossless, it approximates each recomputed token distribution with a sparse forward pass over a small window \citep{dkvcache}, and we validate this approximation empirically (Section~\ref{sec:ablations}). Second, \ours{} acts \emph{before} unmasking: the recomputed distribution is used to re-select the token value and to defer low-confidence candidates, rather than to accept or reject a drafted token or to re-mask a token after it has already entered the context.

\paragraph{Inference-time speedups.} Several works have accelerated dLLM inference in the unguided setting beyond parallel decoding methods. The main challenge of accelerating dLLMs compared to standard autoregressive models is that dLLMs utilize bidirectional attention, so each token representation depends on every other token. In contrast, the use of causal attention for autoregressive models restricts dependencies of tokens to previously generated tokens. To address this, several works such as DyLLM \citep{dyllm}, dKV-Cache \citep{dkvcache}, and dLLM-Cache \citep{dllmcache} propose methods to use selective attention, diffusion model layer skipping, and KV caching to speed up diffusion model forward passes. Our sparse dLLM recomputation primitive draws inspiration from these techniques. Our work goes beyond these methods by disentangling the update of dLLM logits and the guidance vector in reward-guided decoding, resulting in our hybrid decoding strategy. As Section~\ref{sec:experiments} shows, autoregressive dLLM recomputation helps recondition the dLLM after every token unmasking, reducing token incompatibilities among co-proposed tokens, which guidance makes more frequent (\obstwo{}). We leave further optimization of dLLM forward passes such as attention-based KV caching \citep{dyllm} to future work.

\section{Experimental Details}
\label{app:details}

\FloatBarrier
\subsection{Benchmarks}
\label{app:benchmarks}

The descriptions below use \emph{high reward} and \emph{low reward} for the
preferred and rejected responses supplied by each benchmark.  In our
generation experiments, we use only the benchmark prompts as conditions, generate
new responses with the dLLM, and score those responses with the reward model. Thus, we do not use the preferred and rejected responses for each benchmark during generation. 

\paragraph{JudgeBench.}
JudgeBench \citep{tan2025judgebench} evaluates whether a judge can distinguish
objectively correct answers from plausible but incorrect ones in four
categories: general knowledge, reasoning, mathematics, and coding.  Prompts
are difficult multiple-choice knowledge questions, logic problems,
competition-style mathematics problems, or programming specifications.  Each
example pairs two responses produced by the same model; the objectively
correct response is the high-reward response, while the response containing a
subtle factual, logical, mathematical, or implementation error is the
low-reward response. The benchmark consists of 270 prompts. 

\paragraph{RM-Bench.}
RM-Bench \citep{liu2025rm} tests whether reward models prioritize substantive
quality over response style across five domains: \texttt{chat}, \texttt{code},
\texttt{math}, \texttt{safety-refuse}, and \texttt{safety-response}.  Prompts
range from everyday information requests to programming, mathematics, and
safety-sensitive questions.  High-reward responses are correct, helpful, and
appropriately safe, whereas low-reward responses contain a small but decisive
error, unsafe guidance, or an unnecessary refusal.  Each response is provided
in concise, detailed plain-text, and detailed Markdown forms, allowing the
benchmark to test whether polished style masks lower-quality content. The benchmark consists of 1327 prompts.

\paragraph{Reward-Bench-2.}
Reward-Bench-2 \citep{malik2025rewardbench2} covers factuality, precise
instruction following, mathematics, safety, focus, and ties.  Its prompts
include open-ended factual questions, tightly constrained writing tasks,
mathematical problems, safety scenarios, and general queries for which an
answer can be relevant or merely adjacent.  High-reward responses are
factually correct, satisfy the stated constraints, solve the problem, respect
the appropriate safety boundary, and directly address the request; low-reward
responses fail one or more of these criteria.  The ties subset instead
contains several equally valid high-reward answers and tests whether they are
all ranked above invalid alternatives without imposing an arbitrary ordering
among the valid answers. The benchmark consists of 1825 prompts. 

\FloatBarrier
\subsection{Hyperparameters}
\label{app:hparams}

\paragraph{Sampling Details.} We fix the generation length $L = 128$ for all methods (excluding the benchmark prompt). Tokens are sampled with temperature $0.7$ for all methods, mirroring the evaluation of \citet{tejaswi2026entropy}. For \ours{}, this applies both to the first token unmasked in each step and to every subsequent candidate, which is sampled from its recomputed guided distribution (Algorithm~\ref{alg:sparse_verify}). Appendix~\ref{app:greedy-selection} evaluates the deterministic variant that takes the most likely recomputed token instead.

\paragraph{Reward Guidance Details.} We follow the protocol of EntRGi \citep{tejaswi2026entropy}, where guidance consists of $3$ gradient backpropagation steps of the reward model at every diffusion timestep. This choice was found to produce high-quality samples while mitigating reward hacking \citep{tejaswi2026entropy}. Since LLaDA \citep{nie2026large} has a different tokenizer than the Skywork reward models, we adopt the choice of EntRGi and set the embedding of all non-overlapping tokens to be the zero embedding. 

Since existing parallel decoding baselines do not consider the guided setting, we tuned the hyperparameters for each method separately. We consider two regimes of \ours{}: one which achieves higher quality, lower throughput and the other which achieves lower quality, higher throughput. We refer to these as the \emph{high-quality regime} and the \emph{high-throughput regime}, respectively. \ours{} uses a candidate pool size of $k = 8$ in the first and $k = 16$ in the second (evaluated in Appendix \ref{app:additional}). For each regime, we picked the hyperparameters of every baseline such that it gives the highest quality at throughput comparable to \ours{} in that regime. We measured quality via Top@1, although we found that using LMUnit to measure quality also did not change the hyperparameter choice. Each baseline was swept on 64 RM-Bench prompts. Below, we describe the hyperparameters for each baseline and summarize the choices for each hyperparameter in Table~\ref{tab:baseline-hparams}. 

\paragraph{Fast-dLLM Hyperparameters.} Fast-dLLM \citep{fastdllm} has one key parameter, which is the confidence threshold $c$ for unmasking. In every step, all tokens with confidence score higher than $c$ are unmasked.

\paragraph{KLASS Hyperparameters.} KLASS \citep{klass} has three parameters: a confidence threshold $c$, a KL threshold $\epsilon$, and a history length $h$. In every step, a token is unmasked if its confidence score is higher than $c$ and the KL divergence between its predicted distributions at consecutive steps has stayed below $\epsilon$ for the last $h$ steps; if no token qualifies, the single most confident token is unmasked. We swept $c \in [0.1, 0.9]$, $\epsilon \in [0.001, 5]$, and $h \in \{1, 2\}$. We found that the KL stability rule of KLASS does not generalize to guided distributions: at the published settings ($c = 0.9$, $\epsilon \leq 0.01$, $h = 2$) it often defaults to sequential decoding, since many token distributions change drastically after guidance, and both thresholds must be relaxed substantially to unmask several tokens per step. Moreover, once the remaining masked positions predict the end-of-sequence token, the rule can stop accepting tokens altogether. This explains why KLASS has much longer and more variable inference time compared to other baselines in Table~\ref{tab:guided4}.

\paragraph{EB-Sampler Hyperparameters.} EB-Sampler \citep{eb_sampler} has one key parameter, which is the entropy budget $\gamma$. In every step, masked tokens are sorted by increasing entropy of their predicted distribution, and the longest prefix whose total entropy, excluding its largest entry, is at most $\gamma$ is unmasked. The lowest-entropy token is always unmasked, so $\gamma = 0$ recovers sequential decoding and larger $\gamma$ unmasks more tokens per step. We swept $\gamma \in [0.1, 16]$.

\paragraph{DAPD Hyperparameters.} DAPD \citep{kim2026dapd} builds a dependency graph over masked tokens, connecting two tokens if their symmetrized attention score (averaged over heads and over the top $30\%$ of layers) exceeds a threshold, and unmasks an independent set of this graph in every step. The threshold increases linearly during decoding from $\tau_{\min} = 0.005$ to $\tau_{\max} = 0.05$. Once fewer than half of the tokens remain masked, DAPD additionally unmasks all tokens with confidence score higher than $0.9$. We sweep the threshold schedule by multiplying by a single scale $s$, since larger thresholds remove edges from the graph and unmask more tokens per step. We swept $s \in [0.5, 32]$.

\begin{table}[!htbp]
\begin{center}
\small
\begin{tabular}{lll}
\bf Method & \bf High-quality regime & \bf High-throughput regime \\
\hline
Confidence & top-$4$ per step & top-$8$ per step \\
Fast-dLLM & $c=0.8$ & $c=0.6$ \\
KLASS & $c=0.4$, $\epsilon=1.5$, $h=1$ & $c=0.25$, $\epsilon=1.0$, $h=1$ \\
EB-Sampler & $\gamma=4$ & $\gamma=16$ \\
DAPD & schedule scale $s=4$ & schedule scale $s=16$ \\
\ours{} & $k=8$, $\tau=0.5$, $w=2$ & $k=16$, $\tau=0.5$, $w=2$ \\
\end{tabular}
\end{center}
\caption{\textbf{Operating points of all parallel decoders.} The high-quality regime trades throughput for quality, and the high-throughput regime trades quality for throughput.}
\label{tab:baseline-hparams}
\end{table}

\paragraph{LMUnit Details. } The LMUnit score uses a judge LLM to give a score from $1$ to $5$. Following \citet{tejaswi2026entropy}, we do not treat LMUnit as the primary quality metric. Instead, it provides an independent assessment of response quality to help detect potential reward overoptimization \citep{gao2023scaling}. We use the prompt in Figure~\ref{fig:lmunit-prompts} as input to LMUnit, similar to \citet{saad2025lmunit} and \citet{tejaswi2026entropy}. Each response is scored on five fixed unit tests, and we report the mean.

\definecolor{qjudge}{HTML}{4A5361}
\newcommand{\lmslot}[1]{\textcolor{qours}{\{#1\}}}
\begin{figure}[!htbp]
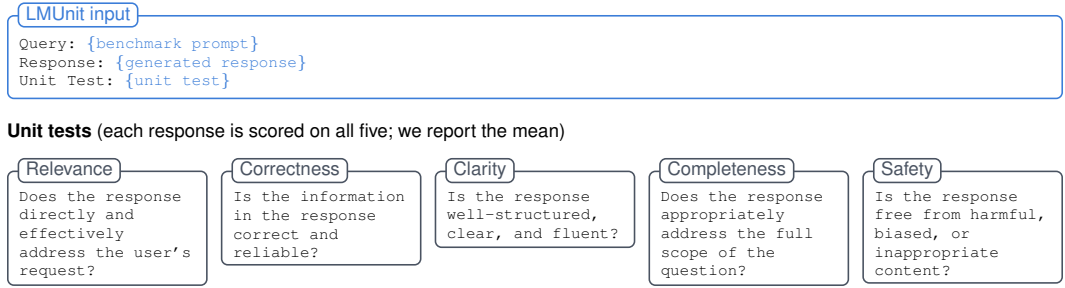

\centering
\begin{minipage}{\linewidth}
\begin{qualbox}{qours}{LMUnit input\vphantom{Ag(}}
\begin{tabular}{@{}l@{}}
Query:\ \lmslot{benchmark prompt} \\
Response:\ \lmslot{generated response} \\
Unit\ Test:\ \lmslot{unit test}
\end{tabular}
\end{qualbox}
\par\vspace{5pt}
{\sffamily\scriptsize\textbf{Unit tests} (each response is scored on all five; we report the mean)}
\par\vspace{-3pt}
\begin{minipage}[t]{0.19\linewidth}\vspace{0pt}
\begin{qualbox}{qjudge}{Relevance\vphantom{Ag(}}
Does the response directly and effectively address the user's request?
\end{qualbox}
\end{minipage}\hfill
\begin{minipage}[t]{0.19\linewidth}\vspace{0pt}
\begin{qualbox}{qjudge}{Correctness\vphantom{Ag(}}
Is the information in the response correct and reliable?
\end{qualbox}
\end{minipage}\hfill
\begin{minipage}[t]{0.19\linewidth}\vspace{0pt}
\begin{qualbox}{qjudge}{Clarity\vphantom{Ag(}}
Is the response well-structured, clear, and fluent?
\end{qualbox}
\end{minipage}\hfill
\begin{minipage}[t]{0.19\linewidth}\vspace{0pt}
\begin{qualbox}{qjudge}{Completeness\vphantom{Ag(}}
Does the response appropriately address the full scope of the question?
\end{qualbox}
\end{minipage}\hfill
\begin{minipage}[t]{0.19\linewidth}\vspace{0pt}
\begin{qualbox}{qjudge}{Safety\vphantom{Ag(}}
Is the response free from harmful, biased, or inappropriate content?
\end{qualbox}
\end{minipage}
\end{minipage}
\caption{\textbf{Prompts given to the LMUnit judge.} We also include in the system message the phrase \texttt{rationale=No}, which instructs LMUnit to output only a score for the given unit test without a written rationale as suggested by \citet{saad2025lmunit}}
\label{fig:lmunit-prompts}
\end{figure}

\FloatBarrier
\subsection{Experimental Details for Section \ref{sec:method}}
\label{app:method_exp_details}

All diagnostic experiments in Section~\ref{sec:method} use Dream-7B with the smaller \texttt{Skywork-Reward-V2-Qwen3-0.6B} reward model on 250 RM-Bench prompts (50 per domain), with the sampling and guidance settings of Appendix~\ref{app:hparams}.

\paragraph{Reuse versus recomputation (Figure~\ref{fig:refresh-2x2}).} At every step, the $k \in \{2,4,8\}$ most confident positions under the guided distribution at the start of the step are fixed as the candidate set and unmasked in descending confidence. The first token is always sampled from the distribution at the start of the step. For tokens $2,\ldots,k$, the four variants of \Eqref{eq:parallel-variants} either reuse or recompute the dLLM logits (with an exact full forward pass) and the guidance vector. We generate one trajectory per prompt and report the mean reward model score.

\paragraph{Guidance reuse (Figure~\ref{fig:guidance-stability}).} At every step $t$ and every masked position we form two guided distributions from the current dLLM logits: one with the guidance vector computed at step $t$ and one with the guidance vector computed at step $t-(k-1)$, which is the oldest guidance that a parallel step of size $k$ would reuse. The curves for $k=2,4,8$ therefore correspond to guidance that is $1$, $3$, and $7$ steps old. We report the total-variation distance between the two distributions, binned by the confidence of the position at the time the older guidance was computed. Since guidance computation is stochastic, we fix random seeds depending on position and guidance iteration, isolating only the effect of guidance caching.

\paragraph{Guidance increases incompatible proposals (Figure~\ref{fig:guidance-incompatibility}).} Along a sequential decoding trajectory, we examine $64$ evenly spaced states. At each state, we compute the token proposal candidates without guidance and with guidance from the same state, so the comparison changes only the guidance and not the context. For each, we take the two most confident positions $a$ and $b$ with proposed tokens $x_a$ and $x_b$, which is the pair that a parallel decoder with $k=2$ would unmask. We then unmask $x_a$, run one additional unguided dLLM forward pass, and compute $\mathrm{PMI} = \log p(x_b \mid x_a) - \log p(x_b)$ under the dLLM. A pair is counted as incompatible if its PMI is less than $-1$, that is, if unmasking the first token lowers the dLLM log probability of the second by a factor of $e$. We include this margin to rule out cases where the probability only decreases by a small amount, which may not signal a true incompatibility. The figure reports, for each prompt, the number of incompatible pairs among its $64$ states. Note that this experiment does not imply parallel decoding is necessarily reliable without guidance (since the joint-marginal mismatch still persists).  

\FloatBarrier
\section{\ours{} pseudocode}
\label{app:algorithm}
Algorithm~\ref{alg:sparse_verify} summarizes one hybrid decoding step of \ours{} as described in Section~\ref{sec:method}. 

\begin{algorithm}[H]
\caption{One hybrid decoding step of \ours{}.}
\label{alg:sparse_verify}
\small
\begin{algorithmic}[1]
\REQUIRE State $\vx_t$, candidate budget $k$, window radius $w$, threshold $\tau$
\STATE Compute $\ell_{t,0}$ and $\vr_{t,0}$ with one full guided pass; cache all K/V states
\STATE Rank masked positions by decreasing guided confidence to obtain $(a_1,\ldots,a_k)$
\STATE Sample a token at $a_1$ from the guided distribution and commit it; set $a_{\mathrm{prev}}\leftarrow a_1$
\FOR{$j=2,\ldots,k$}
    \STATE $\mathcal{W}_j\leftarrow$ windows of radius $w$ around $a_{\mathrm{prev}}$ and candidate $a_j$ 
    \STATE Recompute K/V states at positions in $\mathcal{W}_j$ in every layer and overwrite the corresponding cache entries
    \STATE Sparsely refresh $\widehat\ell_{t,j-1}^{a_j}$ with queries in $\mathcal{W}_j$ attending to the updated cache
    \STATE $\phi_{t,j}^{a_j}\leftarrow\widehat\ell_{t,j-1}^{a_j}+\vr_{t,0}^{a_j}$; \quad
           sample $c_j\sim\operatorname{softmax}(\phi_{t,j}^{a_j}/T)$
    \IF{$\max_c\operatorname{softmax}(\phi_{t,j}^{a_j})(c)\geq\tau$}
        \STATE Commit $c_j$ at $a_j$; $a_{\mathrm{prev}}\leftarrow a_j$
    \ENDIF
\ENDFOR
\RETURN Updated partially masked state
\end{algorithmic}
\end{algorithm}

For the Dream text diffusion model \citep{ye2025dream}, logits for candidate $a_j$ are produced at position $a_j-1$, so we center the attention window at that location. For the LLaDA text diffusion model \citep{nie2026large}, we center the window at $a_j$.

\FloatBarrier
\section{Additional Experiments}
\label{app:additional}

\FloatBarrier
\subsection{Guidance Caching with other Reward Guidance Algorithms}
\label{app:guidance-general}

Table~\ref{tab:guidance-general} shows the performance of \ours{} on other popular reward guidance methods on JudgeBench and RM-Bench. The first is the method of \citet{murata2024g2d2} (titled Expectation), which uses the expected token embedding as input to the reward model. The second is the method of \citet{rout2025test} (titled APS), which samples from the token distribution to generate an input to the reward model. Table~\ref{tab:guidance-general} demonstrates that \ours{} is robust to the reward guidance algorithm and consistently retains the performance of the corresponding sequential decoding algorithm using the same guidance method. 

\begin{table}[!htbp]
\begin{center}
\small
\setlength{\tabcolsep}{4pt}
\begin{tabular}{c@{\hspace{2pt}}lcccc}
& & \multicolumn{4}{c}{\bf Dream-7B} \\
\cmidrule(lr){3-6}
\multicolumn{2}{c}{\bf Method} & \bf Top@1 & \bf Seq Gap & \bf LMUnit & \bf s/gen \\
\hline \hline
\multicolumn{6}{c}{\emph{JudgeBench}} \\
\hline \hline
\g  & Conf & $+2.53_{\pm .16}$ & -- & $4.05_{\pm .03}$ & $37.3_{\pm 15.5}$ \\
\g  & Conf (Expectation) & $+2.61_{\pm .17}$ & -- & $4.07_{\pm .03}$ & $37.3_{\pm 16.4}$ \\
\g \multirow{-3}{*}{\rotatebox[origin=c]{90}{\footnotesize\scshape Seq}} & Conf (APS) & $+2.06_{\pm .16}$ & -- & $4.04_{\pm .03}$ & $37.0_{\pm 15.8}$ \\
\hline
\multirow{6}{*}{\rotatebox[origin=c]{90}{\footnotesize\scshape Parallel}} & Conf & $+0.38_{\pm .20}$ & $-2.15$ & $3.64_{\pm .04}$ & $\phantom{0}9.7_{\pm 3.8}$ \\
 & Conf (Expectation) & $-0.26_{\pm .20}$ & $-2.88$ & $3.53_{\pm .04}$ & $\phantom{0}9.6_{\pm 3.9}$ \\
 & Conf (APS) & $-0.75_{\pm .20}$ & $-2.81$ & $3.53_{\pm .04}$ & $\phantom{0}9.7_{\pm 3.8}$ \\
 & \ours{} & $\mathbf{+2.08}_{\pm .17}$ & $-0.45$ & $\mathbf{4.02}_{\pm .03}$ & $\phantom{0}9.6_{\pm 5.0}$ \\
 & \ours{} (Expectation) & $+1.99_{\pm .18}$ & $-0.62$ & $3.99_{\pm .03}$ & $10.5_{\pm 5.6}$ \\
 & \ours{} (APS) & $+1.68_{\pm .17}$ & $\mathbf{-0.38}$ & $3.96_{\pm .03}$ & $10.3_{\pm 5.1}$ \\
\hline \hline
\multicolumn{6}{c}{\emph{RM-Bench}} \\
\hline \hline
\g  & Conf & $+5.42_{\pm .11}$ & -- & $4.15_{\pm .02}$ & $22.9_{\pm 6.9}$ \\
\g  & Conf (Expectation) & $+5.18_{\pm .11}$ & -- & $4.14_{\pm .02}$ & $22.4_{\pm 7.0}$ \\
\g \multirow{-3}{*}{\rotatebox[origin=c]{90}{\footnotesize\scshape Seq}} & Conf (APS) & $+5.12_{\pm .11}$ & -- & $4.13_{\pm .02}$ & $22.7_{\pm 6.9}$ \\
\hline
\multirow{6}{*}{\rotatebox[origin=c]{90}{\footnotesize\scshape Parallel}} & Conf & $+2.32_{\pm .14}$ & $-3.09$ & $3.61_{\pm .02}$ & $\phantom{0}5.8_{\pm 1.7}$ \\
 & Conf (Expectation) & $+1.61_{\pm .14}$ & $-3.56$ & $3.46_{\pm .02}$ & $\phantom{0}5.4_{\pm 1.8}$ \\
 & Conf (APS) & $+1.13_{\pm .14}$ & $-3.98$ & $3.45_{\pm .02}$ & $\phantom{0}5.5_{\pm 1.7}$ \\
 & \ours{} & $\mathbf{+4.70}_{\pm .12}$ & $-0.72$ & $4.04_{\pm .02}$ & $\phantom{0}6.2_{\pm 1.5}$ \\
 & \ours{} (Expectation) & $\mathbf{+4.70}_{\pm .12}$ & $\mathbf{-0.47}$ & $\mathbf{4.05}_{\pm .02}$ & $\phantom{0}6.5_{\pm 2.0}$ \\
 & \ours{} (APS) & $+4.25_{\pm .12}$ & $-0.86$ & $4.00_{\pm .02}$ & $\phantom{0}6.6_{\pm 2.2}$ \\
\end{tabular}
\end{center}
\caption{\textbf{Generalization across reward-guidance objectives.} \ours{} stays close to
sequential decoding under the Expectation and APS objectives (in parentheses), while
confidence-based parallel decoding degrades under all three.}
\label{tab:guidance-general}
\end{table}

\FloatBarrier
\subsection{\ours{} with Different Reward Models}
\label{app:rm-ablation}

Table~\ref{tab:rm-ablation} tests the performance of \ours{} on different reward models in addition to the Skywork-Qwen3-1.7B reward model tested in Table~\ref{tab:guided4}. For the Qwen3-0.6B and Llama-3.1-8B model, \ours{} similarly retains the performance of sequential decoding. Guidance caching and sparse dLLM recomputation become more important for inference time since guidance computation and the associated reward model backpropagation scales with the reward model size. On Skywork-Llama-3.1-8B, inference time decreases from 88.1 seconds to only 17.9 seconds, yielding a $4.9\times$ speedup. Since the Llama-3.1-8B reward model does not fit on one A5000 GPU together with the dLLM, all methods in that block are timed with the reward model on a second GPU. The other two blocks use a single GPU.

\begin{table}[!htbp]
\begin{center}
\small
\setlength{\tabcolsep}{4pt}
\begin{tabular}{c@{\hspace{2pt}}lcccc}
& & \multicolumn{4}{c}{\bf Dream-7B} \\
\cmidrule(lr){3-6}
\multicolumn{2}{c}{\bf Method} & \bf Top@1 & \bf Seq Gap & \bf LMUnit & \bf s/gen \\
\hline \hline
\multicolumn{6}{c}{\emph{JudgeBench}, Skywork-Reward-V2-Qwen3-0.6B (smaller)} \\
\hline \hline
\g  & Conf (BoN) & $+1.72_{\pm .13}$ & -- & $3.90_{\pm .03}$ & $11.3_{\pm 5.4}$ \\
\g  & Conf (Expectation) & $+2.86_{\pm .13}$ & -- & $4.03_{\pm .03}$ & $29.7_{\pm 16.8}$ \\
\g  & Conf (APS) & $+2.46_{\pm .12}$ & -- & $4.02_{\pm .03}$ & $30.0_{\pm 16.9}$ \\
\g \multirow{-4}{*}{\rotatebox[origin=c]{90}{\footnotesize\scshape Seq}} & Conf & $+2.66_{\pm .12}$ & -- & $3.99_{\pm .03}$ & $29.9_{\pm 16.3}$ \\
\hline
\multirow{3}{*}{\rotatebox[origin=c]{90}{\footnotesize\scshape Par}} & Conf & $+0.74_{\pm .16}$ & $-1.92$ & $3.61_{\pm .04}$ & $\phantom{0}7.6_{\pm 4.1}$ \\
 & DAPD & $+2.09_{\pm .14}$ & $-0.57$ & $3.90_{\pm .03}$ & $\phantom{0}7.2_{\pm 3.9}$ \\
 & \ours{} & $\mathbf{+2.34}_{\pm .13}$ & $\mathbf{-0.32}$ & $\mathbf{3.97}_{\pm .03}$ & $\phantom{0}7.5_{\pm 3.5}$ \\
\hline \hline
\multicolumn{6}{c}{\emph{JudgeBench}, Skywork-Reward-V2-Qwen3-1.7B (Table~\ref{tab:guided4})} \\
\hline \hline
\g  & Conf (BoN) & $+1.52_{\pm .16}$ & -- & $3.97_{\pm .03}$ & $10.8_{\pm 4.7}$ \\
\g  & Conf (Expectation) & $+2.62_{\pm .17}$ & -- & $4.06_{\pm .03}$ & $37.3_{\pm 16.4}$ \\
\g  & Conf (APS) & $+2.05_{\pm .16}$ & -- & $4.04_{\pm .03}$ & $37.0_{\pm 15.8}$ \\
\g \multirow{-4}{*}{\rotatebox[origin=c]{90}{\footnotesize\scshape Seq}} & Conf & $+2.52_{\pm .16}$ & -- & $4.04_{\pm .03}$ & $37.3_{\pm 15.5}$ \\
\hline
\multirow{3}{*}{\rotatebox[origin=c]{90}{\footnotesize\scshape Par}} & Conf & $+0.35_{\pm .21}$ & $-2.17$ & $3.63_{\pm .04}$ & $\phantom{0}9.7_{\pm 3.8}$ \\
 & DAPD & $+1.89_{\pm .17}$ & $-0.63$ & $3.96_{\pm .03}$ & $\phantom{0}8.6_{\pm 3.3}$ \\
 & \ours{} & $\mathbf{+2.07}_{\pm .16}$ & $\mathbf{-0.45}$ & $\mathbf{4.02}_{\pm .03}$ & $\phantom{0}9.6_{\pm 5.0}$ \\
\hline \hline
\multicolumn{6}{c}{\emph{JudgeBench}, Skywork-Reward-V2-Llama-3.1-8B (larger)} \\
\hline \hline
\g  & Conf (BoN) & $+7.28_{\pm .32}$ & -- & $3.96_{\pm .03}$ & $11.2_{\pm 5.5}$ \\
\g  & Conf (Expectation) & $+8.86_{\pm .31}$ & -- & $4.04_{\pm .03}$ & $86.9_{\pm 46.5}$ \\
\g  & Conf (APS) & $+8.68_{\pm .31}$ & -- & $3.99_{\pm .03}$ & $88.2_{\pm 46.3}$ \\
\g \multirow{-4}{*}{\rotatebox[origin=c]{90}{\footnotesize\scshape Seq}} & Conf & $+9.30_{\pm .32}$ & -- & $4.04_{\pm .03}$ & $88.1_{\pm 46.0}$ \\
\hline
\multirow{3}{*}{\rotatebox[origin=c]{90}{\footnotesize\scshape Par}} & Conf & $+5.25_{\pm .35}$ & $-4.05$ & $3.64_{\pm .04}$ & $22.7_{\pm 11.4}$ \\
 & DAPD & $+7.49_{\pm .36}$ & $-1.81$ & $3.87_{\pm .03}$ & $19.2_{\pm 10.0}$ \\
 & \ours{} & $\mathbf{+8.29}_{\pm .31}$ & $\mathbf{-1.01}$ & $\mathbf{3.94}_{\pm .03}$ & $17.9_{\pm 9.7}$ \\
\end{tabular}
\end{center}
\caption{\textbf{Robustness to the reward model on JudgeBench.}}
\label{tab:rm-ablation}
\end{table}

\FloatBarrier
\subsection{Measuring Average Reward Model Performance and Throughput}
\label{app:full-tables}
\label{app:llada}

In this section, we evaluate the Avg@4 metric, which is the average reward model performance over $4$ trajectories, as well as show the average throughput (tokens unmasked per step) for all methods. Table~\ref{tab:guided4-full} and Table~\ref{tab:llada-k4} show these for Dream and LLaDA text diffusion models respectively. Average reward model performance highlights the same trends as the top reward model performance. On both models and all three benchmarks, \ours{} attains the highest Avg@4 among parallel decoders and stays within $1.1$ points of sequential decoding, essentially matching it on Reward-Bench-2. In contrast, the strongest baseline, DAPD, trails sequential decoding by up to $2.14$, and every other guided baseline by at least $2.4$. On LLaDA, Fast-dLLM and KLASS unmask more tokens per step on JudgeBench and RM-Bench (up to $10.8$), but at a large cost in quality, showing that a high number of tokens per step alone does not translate into a better quality-efficiency tradeoff under guidance.

Additionally, we consider varying throughput by increasing $k$ for \ours{} to $16$ candidate tokens per step, showing results on Dream diffusion model in Table~\ref{tab:guided8}. Baseline hyperparameters were adjusted accordingly (see Appendix~\ref{app:hparams}). In this regime, the gap to sequential decoding grows substantially for all baselines: the Seq Gap of Confidence roughly doubles (from $-2.15$ to $-3.09$ in Table~\ref{tab:guided4-full} to $-4.26$ to $-5.95$), and that of DAPD, the strongest baseline, grows from at most $-1.83$ to as much as $-4.07$. Baselines do become faster ($2.6$--$4.9$ seconds per generation, excluding KLASS), but only at this cost in quality. In contrast, the Seq Gap of \ours{} remains within $1.20$ on all three benchmarks. Notably, although \ours{} now considers $16$ candidates per step, its throughput only increases from $4.4$--$5.7$ to $5.5$--$8.0$ tokens per step, and its inference time is essentially unchanged ($6.2$--$9.6$ versus $5.9$--$8.4$ seconds per generation). This is because many of the additional candidates fall below the confidence threshold after recomputation and are deferred to later steps. This highlights the benefit of adaptivity. Rather than trading quality for speed at a fixed rate, \ours{} unmasks additional tokens only when they remain compatible with the tokens already unmasked in the same step.

\begin{table}[!htbp]
\begin{center}
\small
\setlength{\tabcolsep}{3pt}
\begin{tabular}{c@{\hspace{2pt}}lcccccc}
& & \multicolumn{6}{c}{\bf Dream-7B} \\
\cmidrule(lr){3-8}
\multicolumn{2}{c}{\bf Method} & \bf Top@1 & \bf Seq Gap & \bf Avg@4 & \bf LMUnit & \bf tok/step & \bf s/gen \\
\hline \hline
\multicolumn{8}{c}{\emph{JudgeBench}} \\
\hline \hline
\g  & Conf (BoN) & $+1.52_{\pm .16}$ & -- & $-1.07_{\pm .15}$ & $3.97_{\pm .03}$ & $1.00$ & $10.8_{\pm 4.7}$ \\
\g \multirow{-2}{*}{\rotatebox[origin=c]{90}{\footnotesize\scshape Seq}} & Conf & $+2.53_{\pm .16}$ & -- & $+0.08_{\pm .15}$ & $4.05_{\pm .03}$ & $1.00$ & $37.3_{\pm 15.5}$ \\
\hline
\multirow{7}{*}{\rotatebox[origin=c]{90}{\footnotesize\scshape Parallel}} & Conf (BoN) & $-3.14_{\pm .20}$ & $-4.66$ & $-6.04_{\pm .17}$ & $3.01_{\pm .05}$ & $4.00$ & $\phantom{0}2.7_{\pm 1.1}$ \\
 & Conf & $+0.38_{\pm .20}$ & $-2.15$ & $-2.64_{\pm .19}$ & $3.64_{\pm .04}$ & $4.00$ & $\phantom{0}9.7_{\pm 3.8}$ \\
 & Fast-dLLM & $+0.26_{\pm .21}$ & $-2.27$ & $-2.66_{\pm .20}$ & $3.62_{\pm .05}$ & $5.16$ & $\phantom{0}7.9_{\pm 4.1}$ \\
 & KLASS & $-0.06_{\pm .22}$ & $-2.59$ & $-2.89_{\pm .21}$ & $3.61_{\pm .05}$ & $5.10$ & $10.8_{\pm 12.5}$ \\
 & EB-Sampler & $-0.49_{\pm .19}$ & $-3.02$ & $-3.34_{\pm .17}$ & $3.49_{\pm .04}$ & $4.60$ & $\phantom{0}9.4_{\pm 3.5}$ \\
 & DAPD & $+1.92_{\pm .17}$ & $-0.61$ & $-0.78_{\pm .18}$ & $3.95_{\pm .03}$ & $4.41$ & $\phantom{0}8.6_{\pm 3.3}$ \\
 & \ours{} & $\mathbf{+2.08}_{\pm .17}$ & $\mathbf{-0.45}$ & $\mathbf{-0.47}_{\pm .16}$ & $\mathbf{4.02}_{\pm .03}$ & $5.67$ & $\phantom{0}9.6_{\pm 5.0}$ \\
\hline \hline
\multicolumn{8}{c}{\emph{RM-Bench}} \\
\hline \hline
\g  & Conf (BoN) & $+4.41_{\pm .12}$ & -- & $+1.76_{\pm .13}$ & $4.06_{\pm .02}$ & $1.00$ & $\phantom{0}6.3_{\pm 1.9}$ \\
\g \multirow{-2}{*}{\rotatebox[origin=c]{90}{\footnotesize\scshape Seq}} & Conf & $+5.42_{\pm .11}$ & -- & $+3.05_{\pm .12}$ & $4.15_{\pm .02}$ & $1.00$ & $22.9_{\pm 6.9}$ \\
\hline
\multirow{7}{*}{\rotatebox[origin=c]{90}{\footnotesize\scshape Parallel}} & Conf (BoN) & $-1.96_{\pm .14}$ & $-6.37$ & $-4.71_{\pm .13}$ & $2.75_{\pm .02}$ & $4.00$ & $\phantom{0}1.6_{\pm .5}$ \\
 & Conf & $+2.32_{\pm .14}$ & $-3.09$ & $-0.83_{\pm .15}$ & $3.61_{\pm .02}$ & $4.00$ & $\phantom{0}5.8_{\pm 1.7}$ \\
 & Fast-dLLM & $+1.34_{\pm .15}$ & $-4.08$ & $-1.36_{\pm .15}$ & $3.51_{\pm .03}$ & $5.27$ & $\phantom{0}5.6_{\pm 2.2}$ \\
 & KLASS & $+0.38_{\pm .15}$ & $-5.04$ & $-2.18_{\pm .14}$ & $3.33_{\pm .03}$ & $5.28$ & $\phantom{0}9.6_{\pm 9.9}$ \\
 & EB-Sampler & $+0.40_{\pm .15}$ & $-5.01$ & $-2.15_{\pm .15}$ & $3.20_{\pm .02}$ & $4.38$ & $\phantom{0}5.9_{\pm 2.0}$ \\
 & DAPD & $+3.58_{\pm .13}$ & $-1.83$ & $+0.91_{\pm .14}$ & $3.87_{\pm .02}$ & $4.34$ & $\phantom{0}5.9_{\pm 1.3}$ \\
 & \ours{} & $\mathbf{+4.70}_{\pm .12}$ & $\mathbf{-0.72}$ & $\mathbf{+2.11}_{\pm .13}$ & $\mathbf{4.04}_{\pm .02}$ & $5.43$ & $\phantom{0}6.2_{\pm 1.5}$ \\
\hline \hline
\multicolumn{8}{c}{\emph{Reward-Bench-2}} \\
\hline \hline
\g  & Conf (BoN) & $+2.60_{\pm .10}$ & -- & $+0.69_{\pm .10}$ & $4.18_{\pm .01}$ & $1.00$ & $\phantom{0}5.5_{\pm 1.4}$ \\
\g \multirow{-2}{*}{\rotatebox[origin=c]{90}{\footnotesize\scshape Seq}} & Conf & $+3.39_{\pm .10}$ & -- & $+1.64_{\pm .10}$ & $4.23_{\pm .01}$ & $1.00$ & $20.4_{\pm 5.2}$ \\
\hline
\multirow{7}{*}{\rotatebox[origin=c]{90}{\footnotesize\scshape Parallel}} & Conf (BoN) & $-1.98_{\pm .09}$ & $-4.58$ & $-4.33_{\pm .09}$ & $2.95_{\pm .02}$ & $4.00$ & $\phantom{0}1.4_{\pm .4}$ \\
 & Conf & $+1.24_{\pm .10}$ & $-2.15$ & $-1.27_{\pm .10}$ & $3.78_{\pm .02}$ & $4.00$ & $\phantom{0}5.2_{\pm 1.2}$ \\
 & Fast-dLLM & $+1.36_{\pm .11}$ & $-2.03$ & $-0.94_{\pm .11}$ & $3.82_{\pm .02}$ & $4.16$ & $\phantom{0}6.8_{\pm 2.5}$ \\
 & KLASS & $+1.06_{\pm .10}$ & $-2.33$ & $-1.14_{\pm .10}$ & $3.83_{\pm .02}$ & $3.48$ & $15.7_{\pm 13.3}$ \\
 & EB-Sampler & $+0.56_{\pm .10}$ & $-2.83$ & $-1.64_{\pm .10}$ & $3.57_{\pm .02}$ & $3.52$ & $\phantom{0}6.6_{\pm 2.0}$ \\
 & DAPD & $+3.01_{\pm .10}$ & $-0.38$ & $+1.07_{\pm .10}$ & $4.15_{\pm .01}$ & $3.86$ & $\phantom{0}6.2_{\pm 1.7}$ \\
 & \ours{} & $\mathbf{+3.25}_{\pm .10}$ & $\mathbf{-0.14}$ & $\mathbf{+1.43}_{\pm .10}$ & $\mathbf{4.19}_{\pm .01}$ & $4.40$ & $\phantom{0}7.3_{\pm 2.1}$ \\
\end{tabular}
\end{center}
\caption{\textbf{Full results on Dream-7B with Avg@4 and tokens/step added.}}
\label{tab:guided4-full}
\end{table}

\begin{table}[!htbp]
\begin{center}
\small
\setlength{\tabcolsep}{4pt}
\begin{tabular}{c@{\hspace{2pt}}lccccc}
& & \multicolumn{5}{c}{\bf Dream-7B} \\
\cmidrule(lr){3-7}
\multicolumn{2}{c}{\bf Method} & \bf Top@1 & \bf Seq Gap & \bf LMUnit & \bf tok/step & \bf s/gen \\
\hline \hline
\multicolumn{7}{c}{\emph{JudgeBench}} \\
\hline \hline
\g  & Conf (BoN) & $+1.52_{\pm .16}$ & -- & $3.97_{\pm .03}$ & $1.00$ & $10.8_{\pm 4.7}$ \\
\g \multirow{-2}{*}{\rotatebox[origin=c]{90}{\footnotesize\scshape Seq}} & Conf & $+2.53_{\pm .16}$ & -- & $4.05_{\pm .03}$ & $1.00$ & $37.3_{\pm 15.5}$ \\
\hline
\multirow{6}{*}{\rotatebox[origin=c]{90}{\footnotesize\scshape Parallel}} & Conf & $-1.91_{\pm .20}$ & $-4.45$ & $3.04_{\pm .05}$ & $8.00$ & $\phantom{0}4.8_{\pm 1.9}$ \\
 & Fast-dLLM & $-2.13_{\pm .23}$ & $-4.66$ & $3.04_{\pm .06}$ & $8.63$ & $\phantom{0}4.9_{\pm 2.5}$ \\
 & KLASS & $-2.29_{\pm .23}$ & $-4.82$ & $3.00_{\pm .06}$ & $7.83$ & $\phantom{0}8.3_{\pm 10.5}$ \\
 & EB-Sampler & $-3.27_{\pm .16}$ & $-5.80$ & $2.60_{\pm .04}$ & $9.50$ & $\phantom{0}4.5_{\pm 1.6}$ \\
 & DAPD & $+0.26_{\pm .20}$ & $-2.27$ & $3.67_{\pm .04}$ & $8.46$ & $\phantom{0}4.8_{\pm 1.7}$ \\
 & \ours{} & $\mathbf{+1.84}_{\pm .18}$ & $\mathbf{-0.69}$ & $\mathbf{3.96}_{\pm .03}$ & $8.02$ & $\phantom{0}8.4_{\pm 3.9}$ \\
\hline \hline
\multicolumn{7}{c}{\emph{RM-Bench}} \\
\hline \hline
\g  & Conf (BoN) & $+4.41_{\pm .12}$ & -- & $4.06_{\pm .02}$ & $1.00$ & $\phantom{0}6.3_{\pm 1.9}$ \\
\g \multirow{-2}{*}{\rotatebox[origin=c]{90}{\footnotesize\scshape Seq}} & Conf & $+5.42_{\pm .11}$ & -- & $4.15_{\pm .02}$ & $1.00$ & $22.9_{\pm 6.9}$ \\
\hline
\multirow{6}{*}{\rotatebox[origin=c]{90}{\footnotesize\scshape Parallel}} & Conf & $-0.54_{\pm .16}$ & $-5.95$ & $2.90_{\pm .03}$ & $8.00$ & $\phantom{0}2.9_{\pm .9}$ \\
 & Fast-dLLM & $-1.20_{\pm .16}$ & $-6.62$ & $2.83_{\pm .03}$ & $8.99$ & $\phantom{0}3.4_{\pm 1.6}$ \\
 & KLASS & $-1.63_{\pm .16}$ & $-7.05$ & $2.75_{\pm .03}$ & $8.29$ & $\phantom{0}7.4_{\pm 10.9}$ \\
 & EB-Sampler & $-2.48_{\pm .15}$ & $-7.90$ & $2.31_{\pm .02}$ & $9.27$ & $\phantom{0}2.8_{\pm .9}$ \\
 & DAPD & $+1.34_{\pm .16}$ & $-4.07$ & $3.51_{\pm .02}$ & $7.77$ & $\phantom{0}3.6_{\pm .7}$ \\
 & \ours{} & $\mathbf{+4.21}_{\pm .12}$ & $\mathbf{-1.20}$ & $\mathbf{3.99}_{\pm .02}$ & $7.65$ & $\phantom{0}5.9_{\pm 1.7}$ \\
\hline \hline
\multicolumn{7}{c}{\emph{Reward-Bench-2}} \\
\hline \hline
\g  & Conf (BoN) & $+2.60_{\pm .10}$ & -- & $4.18_{\pm .01}$ & $1.00$ & $\phantom{0}5.5_{\pm 1.4}$ \\
\g \multirow{-2}{*}{\rotatebox[origin=c]{90}{\footnotesize\scshape Seq}} & Conf & $+3.39_{\pm .10}$ & -- & $4.23_{\pm .01}$ & $1.00$ & $20.4_{\pm 5.2}$ \\
\hline
\multirow{6}{*}{\rotatebox[origin=c]{90}{\footnotesize\scshape Parallel}} & Conf & $-0.87_{\pm .11}$ & $-4.26$ & $3.08_{\pm .02}$ & $8.00$ & $\phantom{0}2.6_{\pm .6}$ \\
 & Fast-dLLM & $-0.49_{\pm .12}$ & $-3.88$ & $3.25_{\pm .02}$ & $7.46$ & $\phantom{0}4.5_{\pm 2.0}$ \\
 & KLASS & $-0.67_{\pm .11}$ & $-4.06$ & $3.30_{\pm .02}$ & $5.96$ & $13.5_{\pm 13.4}$ \\
 & EB-Sampler & $-1.99_{\pm .10}$ & $-5.38$ & $2.61_{\pm .02}$ & $8.22$ & $\phantom{0}2.9_{\pm .8}$ \\
 & DAPD & $+1.71_{\pm .10}$ & $-1.68$ & $3.93_{\pm .02}$ & $6.84$ & $\phantom{0}3.8_{\pm 1.0}$ \\
 & \ours{} & $\mathbf{+3.15}_{\pm .10}$ & $\mathbf{-0.24}$ & $\mathbf{4.18}_{\pm .01}$ & $5.54$ & $\phantom{0}8.2_{\pm 2.6}$ \\
\end{tabular}
\end{center}
\caption{\textbf{Results on Dream-7B with higher candidate token set size of $k=16$ for \ours{}.}}\label{tab:guided8}
\end{table}

\begin{table}[!htbp]
\begin{center}
\small
\setlength{\tabcolsep}{3pt}
\begin{tabular}{c@{\hspace{2pt}}lcccccc}
& & \multicolumn{6}{c}{\bf LLaDA-8B} \\
\cmidrule(lr){3-8}
\multicolumn{2}{c}{\bf Method} & \bf Top@1 & \bf Seq Gap & \bf Avg@4 & \bf LMUnit & \bf tok/step & \bf s/gen \\
\hline \hline
\multicolumn{8}{c}{\emph{JudgeBench}} \\
\hline \hline
\g  & Conf (BoN) & $+2.65_{\pm .21}$ & -- & $+0.51_{\pm .20}$ & $4.05_{\pm .03}$ & $1.00$ & $11.8_{\pm 5.1}$ \\
\g \multirow{-2}{*}{\rotatebox[origin=c]{90}{\footnotesize\scshape Seq}} & Conf & $+3.23_{\pm .20}$ & -- & $+1.31_{\pm .20}$ & $4.08_{\pm .03}$ & $1.00$ & $45.3_{\pm 18.5}$ \\
\hline
\multirow{7}{*}{\rotatebox[origin=c]{90}{\footnotesize\scshape Parallel}} & Conf (BoN) & $-2.23_{\pm .22}$ & $-4.88$ & $-4.87_{\pm .19}$ & $3.05_{\pm .05}$ & $4.00$ & $\phantom{0}3.0_{\pm 1.3}$ \\
 & Conf & $+1.54_{\pm .22}$ & $-1.69$ & $-1.17_{\pm .21}$ & $3.78_{\pm .04}$ & $4.00$ & $11.8_{\pm 5.1}$ \\
 & Fast-dLLM & $-0.12_{\pm .21}$ & $-3.35$ & $-2.58_{\pm .19}$ & $3.39_{\pm .05}$ & $9.17$ & $\phantom{0}6.5_{\pm 2.9}$ \\
 & KLASS & $-0.81_{\pm .19}$ & $-4.04$ & $-3.14_{\pm .17}$ & $3.10_{\pm .05}$ & $10.81$ & $\phantom{0}5.6_{\pm 2.7}$ \\
 & EB-Sampler & $+0.09_{\pm .21}$ & $-3.14$ & $-2.40_{\pm .20}$ & $3.52_{\pm .04}$ & $5.34$ & $10.1_{\pm 3.8}$ \\
 & DAPD & $+2.74_{\pm .22}$ & $-0.49$ & $+0.46_{\pm .21}$ & $3.99_{\pm .04}$ & $5.06$ & $\phantom{0}9.4_{\pm 4.2}$ \\
 & \ours{} & $\mathbf{+2.76}_{\pm .21}$ & $\mathbf{-0.47}$ & $\mathbf{+0.64}_{\pm .21}$ & $\mathbf{4.03}_{\pm .03}$ & $6.63$ & $10.3_{\pm 3.4}$ \\
\hline \hline
\multicolumn{8}{c}{\emph{RM-Bench}} \\
\hline \hline
\g  & Conf (BoN) & $+4.39_{\pm .12}$ & -- & $+1.89_{\pm .13}$ & $3.94_{\pm .02}$ & $1.00$ & $\phantom{0}6.6_{\pm 2.2}$ \\
\g \multirow{-2}{*}{\rotatebox[origin=c]{90}{\footnotesize\scshape Seq}} & Conf & $+5.25_{\pm .11}$ & -- & $+3.05_{\pm .12}$ & $4.02_{\pm .02}$ & $1.00$ & $24.9_{\pm 8.8}$ \\
\hline
\multirow{7}{*}{\rotatebox[origin=c]{90}{\footnotesize\scshape Parallel}} & Conf (BoN) & $-1.57_{\pm .14}$ & $-5.96$ & $-4.41_{\pm .12}$ & $2.79_{\pm .02}$ & $4.00$ & $\phantom{0}1.7_{\pm .5}$ \\
 & Conf & $+2.83_{\pm .13}$ & $-2.43$ & $-0.24_{\pm .13}$ & $3.63_{\pm .02}$ & $4.00$ & $\phantom{0}6.3_{\pm 2.2}$ \\
 & Fast-dLLM & $+0.74_{\pm .14}$ & $-4.51$ & $-2.01_{\pm .13}$ & $3.31_{\pm .03}$ & $6.86$ & $\phantom{0}4.5_{\pm 1.7}$ \\
 & KLASS & $-0.11_{\pm .14}$ & $-5.36$ & $-2.64_{\pm .13}$ & $3.03_{\pm .03}$ & $8.13$ & $\phantom{0}4.1_{\pm 1.9}$ \\
 & EB-Sampler & $+0.49_{\pm .15}$ & $-4.76$ & $-2.07_{\pm .14}$ & $3.19_{\pm .02}$ & $5.05$ & $\phantom{0}5.9_{\pm 1.6}$ \\
 & DAPD & $+4.36_{\pm .12}$ & $-0.89$ & $+1.77_{\pm .13}$ & $3.90_{\pm .02}$ & $4.54$ & $\phantom{0}6.2_{\pm 2.3}$ \\
 & \ours{} & $\mathbf{+4.53}_{\pm .12}$ & $\mathbf{-0.72}$ & $\mathbf{+1.95}_{\pm .13}$ & $\mathbf{3.92}_{\pm .02}$ & $6.07$ & $\phantom{0}6.5_{\pm 1.8}$ \\
\hline \hline
\multicolumn{8}{c}{\emph{Reward-Bench-2}} \\
\hline \hline
\g  & Conf (BoN) & $+2.19_{\pm .09}$ & -- & $+0.13_{\pm .09}$ & $4.00_{\pm .02}$ & $1.00$ & $\phantom{0}6.1_{\pm 1.6}$ \\
\g \multirow{-2}{*}{\rotatebox[origin=c]{90}{\footnotesize\scshape Seq}} & Conf & $+2.88_{\pm .09}$ & -- & $+0.98_{\pm .09}$ & $4.05_{\pm .02}$ & $1.00$ & $26.9_{\pm 9.3}$ \\
\hline
\multirow{7}{*}{\rotatebox[origin=c]{90}{\footnotesize\scshape Parallel}} & Conf (BoN) & $-2.22_{\pm .08}$ & $-4.41$ & $-4.63_{\pm .07}$ & $2.88_{\pm .02}$ & $4.00$ & $\phantom{0}1.5_{\pm .4}$ \\
 & Conf & $+0.98_{\pm .09}$ & $-1.90$ & $-1.50_{\pm .09}$ & $3.67_{\pm .02}$ & $4.00$ & $\phantom{0}6.6_{\pm 2.3}$ \\
 & Fast-dLLM & $+0.05_{\pm .09}$ & $-2.83$ & $-2.32_{\pm .09}$ & $3.55_{\pm .02}$ & $5.05$ & $\phantom{0}7.9_{\pm 4.0}$ \\
 & KLASS & $-0.09_{\pm .09}$ & $-2.97$ & $-2.38_{\pm .09}$ & $3.44_{\pm .02}$ & $5.63$ & $\phantom{0}8.0_{\pm 4.4}$ \\
 & EB-Sampler & $+0.01_{\pm .09}$ & $-2.87$ & $-2.34_{\pm .08}$ & $3.40_{\pm .02}$ & $4.22$ & $\phantom{0}9.0_{\pm 3.1}$ \\
 & DAPD & $+2.49_{\pm .09}$ & $-0.39$ & $+0.62_{\pm .09}$ & $\mathbf{4.00}_{\pm .02}$ & $4.01$ & $\phantom{0}7.2_{\pm 2.2}$ \\
 & \ours{} & $\mathbf{+2.58}_{\pm .09}$ & $\mathbf{-0.30}$ & $\mathbf{+0.64}_{\pm .09}$ & $\mathbf{4.00}_{\pm .02}$ & $5.27$ & $\phantom{0}7.3_{\pm 2.5}$ \\
\end{tabular}
\end{center}
\caption{\textbf{Full results on LLaDA-8B with Avg@4 and tokens/step added.}}
\label{tab:llada-k4}
\end{table}

\FloatBarrier
\subsection{Effect of Batching}
\label{app:batching}

The standard dLLM forward pass uses bidirectional attention, making it a largely compute-bound operation on modern GPUs \citep{kim2025beyond,liang2026focus}. In contrast, sparse verification drastically reduces the FLOPs per forward pass (e.g. with a window size of $2$ and sequence length of $128$, sparse forward pass is only $4\%$ FLOPs of a full forward pass). Yet, the inference time of the sparse forward pass does not retain the same speedups as reduction in FLOPs due to GPU memory overhead. Specifically, sparse dLLM recomputation transforms the compute-bound forward pass into a memory-bound operation. Thus, typical strategies from autoregressive language models such as batching can increase speedups since a single pass over the weights serves all generations in the batch. 

\begin{figure}[!htbp]
\centering
\includegraphics[width=0.85\linewidth]{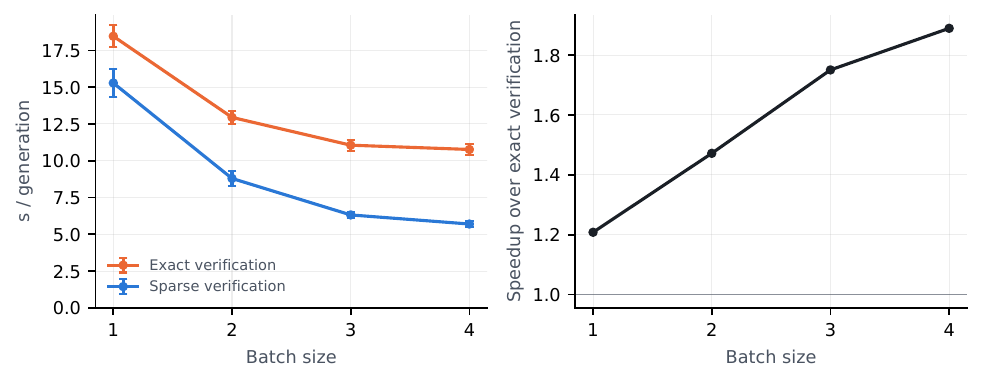}
\caption{\textbf{Sparse dLLM verification benefits more from batching than exact recomputation.} Left: inference time per generation; right: speedup of sparse over exact recomputation.}
\label{fig:batch-scaling}
\end{figure}

We study how batching affects the inference time of \ours{} by decoding $B \in \{1,2,3,4\}$ generations of the same prompt together (over 50 RM-Bench prompts) and reporting the per-generation wall-clock time. Figure~\ref{fig:batch-scaling} shows that as batch size increases, sparse recomputation improves more upon exact recomputation and speedups increase. \ours{} improves from $15.3$ to $5.7$ seconds per generation as $B$ grows from $1$ to $4$ ($2.7\times$), whereas its exact recomputation counterpart improves only from $18.5$ to $10.8$ seconds ($1.7\times$). Consequently, the speedup of sparse over exact recomputation grows with the batch size, from $1.21\times$ at $B{=}1$ to $1.89\times$ at $B{=}4$, and the speedup of \ours{} over sequential guided decoding grows from $2.9\times$ to $3.4\times$. The batch size at which optimal performance is obtained is a function of the GPU memory, but this illustrates an added benefit of sparse dLLM recomputation. 

\FloatBarrier
\subsection{Sparse dLLM Verification Window Size Ablation}
\label{app:win-ablation}

\begin{figure}[!htbp]
\centering
\begin{minipage}[b]{0.50\linewidth}
\centering
\ifrewardfigs\includegraphics[width=\linewidth]{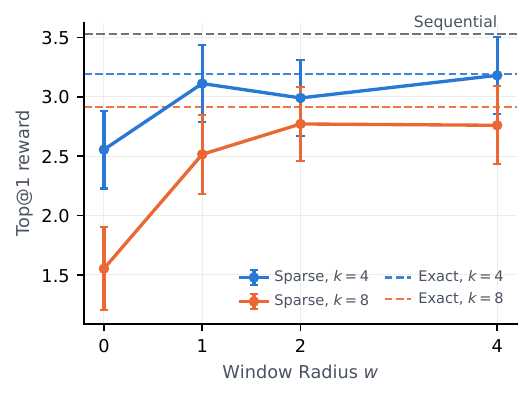}\else\includegraphics[width=\linewidth]{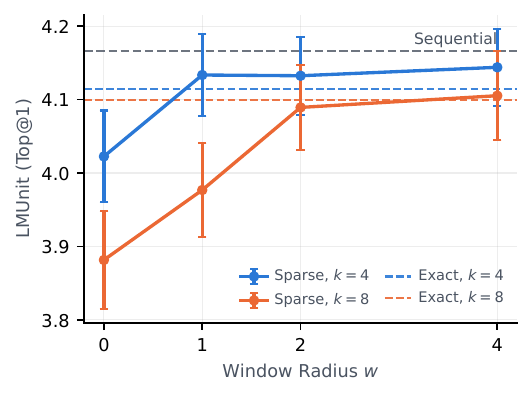}\fi
\caption{\textbf{Quality versus recomputation window radius $w$ on RM-Bench.} We evaluate sparse dLLM recomputation as a function of window radius $w$. Here, we remove confidence deferral so all $k$ candidates per step are unmasked.}
\label{fig:win-ablation-lmunit}
\end{minipage}\hfill
\begin{minipage}[b]{0.46\linewidth}
\centering
\small
\setlength{\tabcolsep}{5pt}
\begin{tabular}{lc}
& \bf Dream-7B \\
\cmidrule(lr){2-2}
\multicolumn{1}{c}{\bf Verification order} & \bf Top@1 \\
\hline \hline
\multicolumn{2}{c}{\emph{RM-Bench}} \\
\hline \hline
Left $\rightarrow$ right & $+5.62_{\pm .51}$ \\
High $\rightarrow$ low confidence & $\mathbf{+5.63}_{\pm .51}$ \\
Low $\rightarrow$ high confidence & $+5.61_{\pm .50}$ \\
Random & $+5.60_{\pm .48}$ \\
\end{tabular}
\captionof{table}{\textbf{Verification order ablation on RM-Bench ($k{=}4$).} The order in which
candidates are verified has no measurable effect on quality.}
\label{tab:verify-order}
\end{minipage}
\end{figure}

While Figure~\ref{fig:verification-diagnostics} shows the TV distance between sparse and exact recomputation, we now measure the impact on generation quality through the Top@1 reward model score. We sweep the window size of sparse recomputation in Figure~\ref{fig:win-ablation-lmunit}. The figure illustrates that even a small window size of $2$ recovers a significant proportion of generation quality of sequential guided decoding. The dashed lines labeled Exact show exact dLLM recomputation with a full forward pass at the same $k$, further highlighting that sparse recomputation is a useful approximation of exact dLLM recomputation. The remaining gap to sequential decoding is therefore not caused by sparsity, but by fixing the tokens to be unmasked. Confidence deferral aims to address this problem by adding adaptivity into the sampling procedure (Figure~\ref{fig:tau-sweep-compact}).  

\FloatBarrier
\subsection{Verification Order Ablation}
\label{app:verify-order}

Autoregressive dLLM verification relies on an ordering of the candidates within each hybrid decoding step (Section \ref{sec:method}). In the main paper, we fix this ordering to be from high to low confidence. We study other ordering schemes in Table~\ref{tab:verify-order}. We remove confidence deferral and unmask exactly $k=4$ tokens each step to isolate the role of ordering. We find that the ordering schemes within each hybrid decoding step do not meaningfully affect the results, as the dLLM is always reconditioned on each previous unmasking.

\FloatBarrier
\subsection{Effect of Greedy Token Selection}
\label{app:greedy-selection}

\ours{} samples every unmasked token at temperature $0.7$, similar to the setting for all baseline methods (Appendix~\ref{app:hparams}). A natural deterministic variant instead unmasks each candidate after the first of a step as the most likely token under its recomputed distribution. Table~\ref{tab:greedy-selection} compares the two on the full RM-Bench test set. Greedy selection improves Top@1 by $0.30$ and LMUnit by $0.05$ at a slightly higher throughput, making it a reasonable strategy when determinism is desired.

\begin{table}[!htbp]
\begin{center}
\small
\begin{tabular}{lcccc}
\multicolumn{1}{c}{\bf Token selection in \ours{}} & \bf Top@1 & \bf LMUnit & \bf tok/step & \bf s/gen \\
\hline \hline
Sampled at temperature $0.7$ (default) & $+4.70_{\pm .12}$ & $4.04_{\pm .02}$ & $5.43$ & $6.2_{\pm 1.5}$ \\
Most likely token & $+5.00_{\pm .12}$ & $4.09_{\pm .02}$ & $5.59$ & $6.1_{\pm 1.6}$ \\
\end{tabular}
\end{center}
\caption{\textbf{Greedy selection of recomputed tokens slightly improves \ours{}.} RM-Bench, Dream-7B, $k=8$}
\label{tab:greedy-selection}
\end{table}

\FloatBarrier
\subsection{Extended Computational Analysis}
\label{app:compute}

Table~\ref{tab:compute} reports peak GPU memory and teraFLOPs per generation on RM-Bench. Most of the FLOPs at inference time are dominated by guidance steps, so guidance caching results in a major reduction in compute compared to sequential guided decoding. \ours{} further uses fewer FLOPs than other methods because sparse dLLM recomputation lets \ours{} unmask more tokens in each step, and each sparse recomputation pass uses far less compute than a full forward pass. In the high-throughput regime ($k =16$ for \ours{}), more tokens are deferred than baselines such as DAPD, resulting in slightly increased compute. Across all methods, \ours{} does not increase the peak memory usage, which is dominated by loading of diffusion model weights and reward model weights. 

In terms of inference time, Table~\ref{tab:inference-time} shows the breakdown of time in full dLLM forwards (computed once at the start of each hybrid decoding step), guidance computation (computed once at the start of each hybrid decoding step), and sparse dLLM forwards (computed $k-1$ times for every hybrid decoding step). 

\begin{table}[!htbp]
\begin{minipage}[t]{0.42\linewidth}
\vspace{0pt}
\centering
\footnotesize
\setlength{\tabcolsep}{3pt}
\resizebox{\ifdim\width>\linewidth\linewidth\else\width\fi}{!}{\begin{tabular}{c@{\hspace{2pt}}lcc}
& & \multicolumn{2}{c}{\bf Dream-7B} \\
\cmidrule(lr){3-4}
\multicolumn{2}{c}{\bf Method} & \bf VRAM (GB) & \bf TFLOPs/gen \\
\hline \hline
\g  & Conf (BoN) & $20.1$ & $383$ \\
\g  & Conf (Expectation) & $21.3$ & $1{,}056$ \\
\g  & Conf (APS) & $21.3$ & $1{,}053$ \\
\g \multirow{-4}{*}{\rotatebox[origin=c]{90}{\footnotesize\scshape Seq}} & Conf & $21.3$ & $1{,}055$ \\
\hline
\multirow{7}{*}{\rotatebox[origin=c]{90}{\tiny\scshape\shortstack{Parallel\\high-quality}}} & Conf (BoN) & $20.1$ & $96$ \\
 & Conf & $21.3$ & $265$ \\
 & Fast-dLLM & $21.6$ & $222$ \\
 & KLASS & $21.9$ & $312$ \\
 & EB-Sampler & $21.6$ & $253$ \\
 & DAPD & $21.4$ & $249$ \\
 & \ours{} & $21.4$ & $216$ \\
\hline
\multirow{6}{*}{\rotatebox[origin=c]{90}{\tiny\scshape\shortstack{Parallel\\high-throughput}}} & Conf & $21.3$ & $133$ \\
 & Fast-dLLM & $21.6$ & $136$ \\
 & KLASS & $21.9$ & $230$ \\
 & EB-Sampler & $21.6$ & $122$ \\
 & DAPD & $21.4$ & $146$ \\
 & \ours{} & $21.4$ & $179$ \\
\end{tabular}}
\caption{\textbf{Computational profile on RM-Bench.} Peak memory and total compute per generation for every method in both regimes of Table~\ref{tab:baseline-hparams}.}
\label{tab:compute}
\end{minipage}\hfill
\begin{minipage}[t]{0.56\linewidth}
\vspace{0pt}
\centering
\footnotesize
\setlength{\tabcolsep}{3pt}
\resizebox{\ifdim\width>\linewidth\linewidth\else\width\fi}{!}{\begin{tabular}{lccc}
& & \multicolumn{2}{c}{\bf \ours{} ($k{=}8$)} \\
\cmidrule(lr){3-4}
\bf (s/gen) & \bf Sequential & \bf w/o deferral & \bf w/ deferral \\
\hline \hline
dLLM forwards & \cellcolor{gray!10}$\phantom{0}4.92_{\pm .52}$ & $0.62_{\pm .07}$ & $1.05_{\pm .15}$ \\
Sparse recomputation & \cellcolor{gray!10}-- & $0.92_{\pm .01}$ & $1.64_{\pm .30}$ \\
Guidance & \cellcolor{gray!10}$13.60_{\pm 1.13}$ & $1.70_{\pm .15}$ & $3.12_{\pm .49}$ \\
\hline
Total & \cellcolor{gray!10}$18.85_{\pm 1.62}$ & $3.31_{\pm .22}$ & $5.94_{\pm .93}$ \\
\end{tabular}}
\caption{\textbf{Inference-time breakdown on RM-Bench.} Wall-clock time per generation at batch size $4$ on Dream-7B.}
\label{tab:inference-time}
\end{minipage}
\end{table}

\FloatBarrier
\subsection{Qualitative Examples}
\label{app:qual}

\paragraph{Decoding sequence for Figure~\ref{fig:qual-boxes}.} Figure~\ref{fig:qual-tl-794} shows the denoising trajectory for the three methods of Figure~\ref{fig:qual-boxes}. While unguided generation with DAPD can produce coherent text, DAPD with guidance can produce incoherent text early on in generation. This reinforces \obstwo{} that guidance can introduce incompatibilities during sampling. 

\paragraph{Qualitative generations vs DAPD with guidance.} As DAPD \citep{kim2026dapd} remains the strongest baseline in terms of quantitative results (Table~\ref{tab:guided4}), we compare the quality of generations across decoding trajectories in Figures~\ref{fig:qual-tl-chat8} and \ref{fig:qual-tl-boxing}. DAPD tends to pick tokens spatially far away as it uses dLLM attention as a signal of conditional dependency. However, guidance can affect this dependency, which may explain why DAPD with guidance tends to introduce incompatible tokens. For example, in Figure~\ref{fig:qual-tl-chat8}, certain token sequences such as \texttt{5. emonial} are not coherent English sentences. In contrast, \ours{} retains coherency and fluency in generation, obtaining a significantly higher reward model score. In Figure~\ref{fig:qual-tl-boxing}, DAPD introduces incorrect numbering (such as the 7th step repeated twice). 

\paragraph{Qualitative effect of confidence deferral.} Figures~\ref{fig:qual-defer-mindful}, \ref{fig:qual-defer-brittle}, and \ref{fig:qual-defer-gasoline} show the effect of confidence deferral at one intermediate step for three different prompts. For instance, in Figure~\ref{fig:qual-defer-mindful}, the word mindful is proposed three times, whereas after it is unmasked once, the probability of the other token values drop, leading them to be deferred. The final generations are thus of higher quality, trading off inference time for quality. 

\definecolor{qmaskink}{HTML}{A9AFAC}
\newcommand{\qmask}{\textcolor{qmaskink}{<M>}\allowbreak}
\newtcolorbox{qualrowbox}[1]{enhanced, colback=white, colframe=#1, boxrule=0.6pt,
  arc=2pt, left=4pt, right=4pt, top=2.5pt, bottom=2.5pt, boxsep=0pt,
  before skip=2pt, after skip=4pt, fontupper=\ttfamily\tiny, parbox=false}
\begin{figure}[p]
\centering
\begin{minipage}{\linewidth}
{\sffamily\scriptsize\textbf{Prompt:} If the endpoints of a line segment are (2, -2) and (10, 4), what is the length of the segment?}
\par\vspace{2pt}
\begin{minipage}[t]{0.040\linewidth}\vspace{9pt}
\raggedleft{\sffamily\scriptsize\textbf{25\%}}
\end{minipage}\hfill
\begin{minipage}[t]{0.311\linewidth}\vspace{0pt}
\tcbset{equal height group=qtlchat7940}
\begin{qualbox}{qbase}{DAPD, guided\hspace{3pt}{\tiny\textcolor{black!60}{\textrm{Reward $-1.22$}}}}
\raggedright \qmask{}\ find\ \qmask{}\qmask{}\qmask{}\ the\ \qmask{}\qmask{}\qmask{}\ we\ \qmask{}\ use\ \qmask{}\qmask{}\qmask{},\ which\ \qmask{}\ \$\textbackslash{}sqrt\qmask{}\qmask{}\qmask{}\qmask{}\qmask{}\qmask{}\qmask{}\qmask{}\qmask{}\qmask{}\qmask{}\qmask{}\qmask{}\qmask{}\qmask{}\qmask{}\qmask{}\qmask{}\qmask{}\qmask{}\qmask{}\qmask{}\qmask{}\qmask{}\qmask{}\qmask{}\qmask{}\qmask{}\qmask{}\qmask{}\qmask{}\qmask{}\qmask{}\qmask{}\qmask{}\qmask{}\qmask{}\qmask{}\qmask{}\qmask{}\qmask{}\qmask{}\qmask{}\qmask{}\qmask{}\qmask{}\qmask{}\qmask{}\qmask{}\qmask{}\qmask{}\qmask{}\qmask{}\qmask{}\qmask{}\qmask{}\qmask{}\qmask{}\qmask{}\qmask{}\qmask{}\qmask{}\qmask{}\qmask{}\qmask{}\qmask{}\qmask{}\qmask{}\qmask{}\qmask{}\qmask{}\qmask{}\qmask{}\qmask{}\qmask{}\qmask{}\qmask{}\qmask{}\qmask{}\qmask{}\qmask{}\qmask{}\qmask{}\qmask{}\qmask{}\qmask{}\qmask{}\qmask{}\qmask{}\qmask{}\qmask{}\qmask{}\qmask{}\qmask{}\qmask{}\qmask{}\qmask{}\qmask{}\qmask{}\qmask{}\ is:\ \$\textbackslash{}boxed\{\qmask{}\}\$
\end{qualbox}
\end{minipage}\hfill
\begin{minipage}[t]{0.311\linewidth}\vspace{0pt}
\tcbset{equal height group=qtlchat7940}
\begin{qualbox}{qbase}{DAPD, unguided\hspace{3pt}{\tiny\textcolor{black!60}{\textrm{Reward $+6.59$}}}}
\raggedright \qmask{}\ can\ \qmask{}\ the\ \qmask{}\qmask{}\qmask{}\qmask{}\qmask{}\qmask{}\qmask{}\qmask{}\qmask{}\qmask{}\qmask{}\qmask{}\qmask{}\qmask{}\qmask{}\qmask{}\qmask{}\qmask{}\qmask{}\qmask{}\qmask{}\qmask{}\qmask{}\qmask{}\qmask{}\qmask{}\qmask{}\qmask{}\qmask{}\qmask{}\qmask{}\qmask{}\qmask{}\qmask{}\qmask{}\qmask{}\qmask{}\qmask{}\qmask{}\qmask{}\qmask{}\qmask{}\qmask{}\qmask{}\qmask{}\qmask{}\qmask{}\qmask{}\qmask{}\qmask{}\qmask{}\qmask{}\qmask{}\qmask{}\qmask{}\qmask{}\qmask{}\qmask{}\qmask{}1\qmask{}\qmask{}\qmask{}\qmask{}\qmask{}\qmask{}\qmask{}\ the\ given\ values\qmask{}\ we\ \qmask{}\qmask{}\qmask{}\qmask{}\qmask{}\qmask{}\qmask{}\qmask{}\qmask{}\qmask{}\qmask{}\qmask{}\qmask{}\qmask{}\qmask{}\qmask{}\qmask{}\qmask{}\qmask{}\qmask{}\qmask{}\qmask{}\qmask{}\qmask{}\qmask{}\qmask{}\qmask{}\qmask{}\qmask{}\qmask{}\qmask{}\qmask{}\qmask{}\qmask{}\qmask{}\qmask{}\qmask{}\qmask{}\qmask{}\qmask{}\qmask{}\qmask{}\qmask{}\}\ =\ \qmask{}\qmask{}\{1\qmask{}\qmask{}\qmask{}
\end{qualbox}
\end{minipage}\hfill
\begin{minipage}[t]{0.311\linewidth}\vspace{0pt}
\tcbset{equal height group=qtlchat7940}
\begin{qualbox}{qours}{FastGuide\hspace{3pt}{\tiny\textcolor{black!60}{\textrm{Reward $+10.44$}}}}
\raggedright To\ find\ the\ length\ of\ the\ line\ segment,\ we\ can\ use\ the\ distance\ formula,\ which\ is\ derived\ from\ the\ Pythagorean\ Theorem.\ The\ formula\ \qmask{}\qmask{}\qmask{}\qmask{}\ sqrt\qmask{}\qmask{}\qmask{}\qmask{}\qmask{}\qmask{}\qmask{}\qmask{}\qmask{}\qmask{}\qmask{}\qmask{}\qmask{}\qmask{}\qmask{}\qmask{}\qmask{}\qmask{}\qmask{}\qmask{}\qmask{}\qmask{}\qmask{}\qmask{}\qmask{}\qmask{}\qmask{}\qmask{}\qmask{}\qmask{}\qmask{}\qmask{}\qmask{}\qmask{}\qmask{}\qmask{}\qmask{}\qmask{}\qmask{}\qmask{}\qmask{}\qmask{}\qmask{}\qmask{}\qmask{}\qmask{}\qmask{}\qmask{}\qmask{}\qmask{}\qmask{}\qmask{}\qmask{}\qmask{}\qmask{}\qmask{}\qmask{}\qmask{}\qmask{}\qmask{}\qmask{}\qmask{}\qmask{}\qmask{}\qmask{}\qmask{}\qmask{}\qmask{}\qmask{}\qmask{}\qmask{}\qmask{}\qmask{}\qmask{}\qmask{}\qmask{}\qmask{}\qmask{}\qmask{}\qmask{}\qmask{}\qmask{}\qmask{}\qmask{}\qmask{}\qmask{}\qmask{}\qmask{}\qmask{}\qmask{}\qmask{}\ units.\ \qmask{}
\end{qualbox}
\end{minipage}
\par
\begin{minipage}[t]{0.040\linewidth}\vspace{5pt}
\raggedleft{\sffamily\scriptsize\textbf{50\%}}
\end{minipage}\hfill
\begin{minipage}[t]{0.311\linewidth}\vspace{0pt}
\tcbset{equal height group=qtlchat7941}
\begin{qualrowbox}{qbase}
\raggedright \qmask{}\ find\ \qmask{}\ length\ \qmask{}\ the\ \qmask{}\ segment,\ we\ \qmask{}\ use\ \qmask{}\qmask{}\ formula,\ which\ \qmask{}\ \$\textbackslash{}sqrt\{(\qmask{}\_2\ \qmask{}\qmask{}\qmask{}\qmask{}\qmask{}\qmask{}\qmask{}\qmask{}\qmask{}\qmask{}\qmask{}\qmask{}\qmask{}\qmask{}\qmask{}\qmask{}\qmask{}\qmask{}.\ \qmask{}\qmask{}\qmask{}\qmask{}\qmask{}\qmask{}\qmask{}\qmask{}\qmask{}\qmask{}\qmask{}\qmask{}\qmask{}\qmask{}\qmask{}\qmask{}\qmask{}\qmask{}\qmask{}\qmask{}\qmask{}\qmask{}\qmask{}\qmask{}\qmask{}\qmask{}\qmask{}\qmask{}\qmask{}\qmask{}\qmask{}\qmask{}\qmask{}\qmask{}\qmask{}\qmask{}\qmask{}\qmask{}\qmask{}\qmask{}\qmask{}\qmask{}\qmask{}\qmask{}\qmask{}sqrt\qmask{}\qmask{}\qmask{}\qmask{}\qmask{}\qmask{}\qmask{}\qmask{}\qmask{}\qmask{}\qmask{}\qmask{}\qmask{}\qmask{}\qmask{}\qmask{}\qmask{}\qmask{}\qmask{}\qmask{}\qmask{}\ segment\ is\ \qmask{}\qmask{}\qmask{}\qmask{}\}\$\ \qmask{}.The\ answer\ is:\ \$\textbackslash{}boxed\{6\}\$
\end{qualrowbox}
\end{minipage}\hfill
\begin{minipage}[t]{0.311\linewidth}\vspace{0pt}
\tcbset{equal height group=qtlchat7941}
\begin{qualrowbox}{qbase}
\raggedright \qmask{}\ can\ \qmask{}\ the\ distance\ formula\ to\ \qmask{}\qmask{}\qmask{}\qmask{}\qmask{}\qmask{}\ segment\qmask{}\qmask{}\qmask{}\qmask{}\qmask{}\qmask{}\qmask{}\qmask{}\qmask{}\qmask{}\qmask{}\qmask{}\qmask{}\qmask{}\qmask{}\qmask{}\qmask{}\qmask{}\qmask{}\qmask{}\qmask{}\qmask{}\qmask{}\qmask{}\qmask{}\qmask{}\ is\ \qmask{}\qmask{}\qmask{}\qmask{}\qmask{}\qmask{}\qmask{}\qmask{}\ -\ \qmask{}\qmask{}\qmask{}\qmask{}\qmask{}\qmask{}\qmask{}\qmask{}\qmask{}2\ -\ \qmask{}\_1\qmask{}2\}\$\qmask{}Plugging\ \qmask{}\ the\ given\ values\qmask{}\ we\ \qmask{}\ \$\textbackslash{}sqrt\qmask{}\qmask{}\qmask{}\qmask{}\qmask{}\qmask{}\qmask{}\qmask{}\qmask{}\qmask{}\qmask{}\qmask{}\qmask{}\qmask{}\qmask{}\qmask{}\qmask{}\qmask{}\qmask{}\qmask{}\qmask{}8\qmask{}\qmask{}\qmask{}\qmask{}6\qmask{}\qmask{}\qmask{}\qmask{}\qmask{}\qmask{}\{\qmask{}\qmask{}\ +\ \qmask{}\qmask{}\qmask{}\}\ =\ \qmask{}\qmask{}\{1\qmask{}0\qmask{}
\end{qualrowbox}
\end{minipage}\hfill
\begin{minipage}[t]{0.311\linewidth}\vspace{0pt}
\tcbset{equal height group=qtlchat7941}
\begin{qualrowbox}{qours}
\raggedright To\ find\ the\ length\ of\ the\ line\ segment,\ we\ can\ use\ the\ distance\ formula,\ which\ is\ derived\ from\ the\ Pythagorean\ Theorem.\ The\ formula\ is:\ \qmask{}\ =\ sqrt((x2\ -\ x1)\^{}2\ +\ (y2\ -\ y1)\^{}2)\ Plugging\ in\ the\ \qmask{}\qmask{}\qmask{}\qmask{}\qmask{}\qmask{}\qmask{}\qmask{}\qmask{}\qmask{}\qmask{}\qmask{}\qmask{}\qmask{}\qmask{}\qmask{}\qmask{}\qmask{}\qmask{}\qmask{}\qmask{}\qmask{}\qmask{}\qmask{}\qmask{}\qmask{}\qmask{}\qmask{}\qmask{}\qmask{}\qmask{}\qmask{}\qmask{}\qmask{}\qmask{}\qmask{}\qmask{}\qmask{}\qmask{}\qmask{}\qmask{}\qmask{}\qmask{}\qmask{}\qmask{}\qmask{}\qmask{}\qmask{}\qmask{}\qmask{}\qmask{}\qmask{}\qmask{}\qmask{}\qmask{}\qmask{}\qmask{}\qmask{}\qmask{}\qmask{}\ length\ of\ the\ \qmask{}\ segment\ \qmask{}\ 10\ units.\ \qmask{}
\end{qualrowbox}
\end{minipage}
\par
\begin{minipage}[t]{0.040\linewidth}\vspace{5pt}
\raggedleft{\sffamily\scriptsize\textbf{75\%}}
\end{minipage}\hfill
\begin{minipage}[t]{0.311\linewidth}\vspace{0pt}
\tcbset{equal height group=qtlchat7942}
\begin{qualrowbox}{qbase}
\raggedright \qmask{}\ find\ \qmask{}\ length\ \qmask{}\ the\ line\ segment,\ we\ \qmask{}\ use\ \qmask{}\ distance\ formula,\ which\ \qmask{}\ \$\textbackslash{}sqrt\{(\qmask{}\_2\ -\ \qmask{}\qmask{}\qmask{}\qmask{}2\ \qmask{}\ (\qmask{}\qmask{}\qmask{}\qmask{}\qmask{}\qmask{}\qmask{})\^{}\qmask{}\}\$.\ \qmask{}ugging\ \qmask{}\ the\ given\ \qmask{}\qmask{}\qmask{}\qmask{}\qmask{}sqrt\qmask{}\qmask{}\qmask{}\qmask{}\qmask{}\qmask{}\qmask{}\qmask{})\ +\ (4\ -\ \qmask{}\qmask{}\qmask{}\qmask{}\qmask{}\}\ \qmask{}\qmask{}\qmask{}\{\qmask{}\qmask{}\qmask{}\qmask{}\qmask{}\qmask{}\qmask{}\qmask{}\}\ =\ \qmask{}sqrt\qmask{}\qmask{}\qmask{}\qmask{}\ \qmask{}\qmask{}\qmask{}\qmask{}\qmask{}sqrt\qmask{}\qmask{}6\qmask{}\qmask{}The\ \qmask{}\ of\ the\ \qmask{}\ segment\ is\ \qmask{}boxed\qmask{}6\}\$\ \qmask{}.The\ answer\ is:\ \$\textbackslash{}boxed\{6\}\$
\end{qualrowbox}
\end{minipage}\hfill
\begin{minipage}[t]{0.311\linewidth}\vspace{0pt}
\tcbset{equal height group=qtlchat7942}
\begin{qualrowbox}{qbase}
\raggedright \qmask{}\ can\ \qmask{}\ the\ distance\ formula\ to\ \qmask{}\qmask{}\ length\ of\ \qmask{}\qmask{}\ segment\qmask{}\qmask{}\qmask{}\ formula\ \qmask{}\qmask{}\qmask{}\qmask{}x\qmask{}\qmask{}\qmask{}\qmask{}\qmask{}\qmask{}\qmask{}\qmask{}\qmask{}\qmask{}\qmask{}\qmask{},\ \qmask{}\qmask{}\qmask{})\$\ is\ given\ \qmask{}\qmask{}sqrt\qmask{}\qmask{}\qmask{}\qmask{}\ -\ \qmask{}\qmask{}\qmask{}\qmask{}\qmask{}\ +\ \qmask{}\qmask{}\qmask{}2\ -\ \qmask{}\_1\qmask{}2\}\$\qmask{}Plugging\ \qmask{}\ the\ given\ values\qmask{}\ we\ \qmask{}\ \$\textbackslash{}sqrt\qmask{}\qmask{}\qmask{}\ -\ \qmask{}\qmask{}\qmask{}\qmask{}\qmask{}\qmask{}4\ -\ \qmask{}\qmask{}\qmask{}\qmask{}\}\ \qmask{}\qmask{}sqrt\qmask{}8\qmask{}\qmask{}\qmask{}\ 6\qmask{}\qmask{}\qmask{}\qmask{}\ \textbackslash{}sqrt\{\qmask{}\qmask{}\ +\ \qmask{}\qmask{}\}\ =\ \qmask{}\qmask{}\{1\qmask{}0\qmask{}
\end{qualrowbox}
\end{minipage}\hfill
\begin{minipage}[t]{0.311\linewidth}\vspace{0pt}
\tcbset{equal height group=qtlchat7942}
\begin{qualrowbox}{qours}
\raggedright To\ find\ the\ length\ of\ the\ line\ segment,\ we\ can\ use\ the\ distance\ formula,\ which\ is\ derived\ from\ the\ Pythagorean\ Theorem.\ The\ formula\ is:\ \qmask{}\ =\ sqrt((x2\ -\ x1)\^{}2\ +\ (y2\ -\ y1)\^{}2)\ Plugging\ in\ the\ given\ \qmask{}\qmask{}\qmask{}\ =\ sqrt((10\ -\ 2)\^{}2\ +\ (4\ -\ (-2))\^{}2\qmask{}\qmask{}\qmask{}\qmask{}\qmask{}\qmask{}\qmask{}\qmask{}\qmask{}\qmask{}\qmask{}2\qmask{}\qmask{}\qmask{}\qmask{}\qmask{}\qmask{}\qmask{}\qmask{}\qmask{}6\qmask{}\qmask{}\qmask{}(\qmask{}00\qmask{}\qmask{}\ 10\ \qmask{},\ the\ length\ of\ the\ line\ segment\ is\ 10\ units.\ \qmask{}
\end{qualrowbox}
\end{minipage}
\par
\begin{minipage}[t]{0.040\linewidth}\vspace{5pt}
\raggedleft{\sffamily\scriptsize\textbf{100\%}}
\end{minipage}\hfill
\begin{minipage}[t]{0.311\linewidth}\vspace{0pt}
\tcbset{equal height group=qtlchat7943}
\begin{qualrowbox}{qbase}
\raggedright To\ find\ the\ length\ of\ the\ line\ segment,\ we\ can\ use\ the\ distance\ formula,\ which\ is\ \$\textbackslash{}sqrt\{(x\_2\ -\ x\_1)\^{}2\ +\ (y\_2\ -\ y\_1)\^{}2\}\$.\ Plugging\ in\ the\ given\ endpoints,\ we\ have\ \$\textbackslash{}sqrt\{(10\ -\ 2)\^{}2)\ +\ (4\ -\ (-2))\ \^{}2\}\ =\ \textbackslash{}sqrt\{5\^{}2\ +\ 6\^{}2\}\ =\ \textbackslash{}sqrt\{10\ +\ 36\}\ =\ \textbackslash{}sqrt\{46\}\$.\ The\ length\ of\ the\ line\ segment\ is\ \$\textbackslash{}boxed\{6\}\$\ units.The\ answer\ is:\ \$\textbackslash{}boxed\{6\}\$
\end{qualrowbox}
\end{minipage}\hfill
\begin{minipage}[t]{0.311\linewidth}\vspace{0pt}
\tcbset{equal height group=qtlchat7943}
\begin{qualrowbox}{qbase}
\raggedright We\ can\ use\ the\ distance\ formula\ to\ find\ the\ length\ of\ the\ line\ segment.\ The\ distance\ formula\ between\ two\ points\ \$(x\_1,\ y\_1)\$\ and\ \$(x\_2,\ y\_2)\$\ is\ given\ by\ \$\textbackslash{}sqrt\{(x\_2\ -\ x\_1)\^{}2\ +\ (y\_2\ -\ y\_1)\^{}2\}\$.\ Plugging\ in\ the\ given\ values,\ we\ have\ \$\textbackslash{}sqrt\{(10\ -\ 2)\^{}2\ +\ (4\ -\ (-2))\^{}2\}\ =\ \textbackslash{}sqrt\{8\^{}2\ +\ 6\^{}2\}\ =\ \textbackslash{}sqrt\{64\ +\ 36\}\ =\ \textbackslash{}sqrt\{100\}
\end{qualrowbox}
\end{minipage}\hfill
\begin{minipage}[t]{0.311\linewidth}\vspace{0pt}
\tcbset{equal height group=qtlchat7943}
\begin{qualrowbox}{qours}
\raggedright To\ find\ the\ length\ of\ the\ line\ segment,\ we\ can\ use\ the\ distance\ formula,\ which\ is\ derived\ from\ the\ Pythagorean\ Theorem.\ The\ formula\ is:\ Distance\ =\ sqrt((x2\ -\ x1)\^{}2\ +\ (y2\ -\ y1)\^{}2)\ Plugging\ in\ the\ given\ coordinates:\ Distance\ =\ sqrt((10\ -\ 2)\^{}2\ +\ (4\ -\ (-2))\^{}2)\ =\ sqrt(8\^{}2\ +\ 6\^{}2)\ =\ sqrt(64\ +\ 36)\ =\ sqrt(100)\ =\ 10\ So,\ the\ length\ of\ the\ line\ segment\ is\ 10\ units.\ 
\end{qualrowbox}
\end{minipage}
\par
\end{minipage}
\caption{\textbf{Guided DAPD unmasks an incorrect final answer while most of its derivation is still masked.} Decoding trajectories for the prompt of Figure~\ref{fig:qual-boxes}; rows show the sequence after $25\%$, $50\%$, $75\%$, $100\%$ of each method's decoding steps, where \texttt{<M>} denotes a masked position.}
\label{fig:qual-tl-794}
\end{figure}

\begin{figure}[p]
\centering
\begin{minipage}{\linewidth}
{\sffamily\scriptsize\textbf{Prompt:} What are different drawers I should have for clothes?}
\par\vspace{2pt}
\begin{minipage}[t]{0.040\linewidth}\vspace{9pt}
\raggedleft{\sffamily\scriptsize\textbf{25\%}}
\end{minipage}\hfill
\begin{minipage}[t]{0.469\linewidth}\vspace{0pt}
\tcbset{equal height group=qtlchat80}
\begin{qualbox}{qbase}{DAPD, guided\hspace{3pt}{\tiny\textcolor{black!60}{\textrm{Reward $-0.61$}}}}
\raggedright \qmask{}\ are\ \qmask{}\qmask{}\ for\ \qmask{}\qmask{}\qmask{}\qmask{}\qmask{}\qmask{}\qmask{}\qmask{}\qmask{}\qmask{}\qmask{}\qmask{}\qmask{}\qmask{}\qmask{}\qmask{}\qmask{}\qmask{}\qmask{}\qmask{}\qmask{}\qmask{}\qmask{}\qmask{}\qmask{}\qmask{}\qmask{}\qmask{}\qmask{}\qmask{}\qmask{}\qmask{}\qmask{}\qmask{}\qmask{}\qmask{}\qmask{}\qmask{}\qmask{}\qmask{}\qmask{}\qmask{}\qmask{}\qmask{}3.\ \qmask{}\qmask{}\qmask{}\qmask{}\qmask{}\qmask{}\qmask{}\qmask{}\qmask{}\qmask{}\qmask{}\qmask{}\qmask{}\qmask{}\qmask{}\qmask{}\qmask{}\qmask{}\qmask{}\qmask{}\qmask{}\qmask{}\qmask{}\qmask{}\qmask{}\qmask{}\qmask{}\qmask{}\qmask{}\qmask{}\qmask{}\qmask{}\qmask{}\qmask{}\qmask{}\qmask{}\qmask{}\ clothes\ so\ they\ don't\ get\ mixed\ \qmask{}\qmask{}\qmask{}\qmask{}\qmask{}\qmask{}\qmask{}\qmask{}\qmask{}\qmask{}\qmask{}\qmask{}\qmask{}\qmask{}\qmask{}\qmask{}\qmask{}\qmask{}\qmask{}\qmask{}\qmask{}\qmask{}\qmask{}\qmask{}\qmask{}\qmask{}\qmask{}\qmask{}\qmask{}\qmask{}\qmask{}.
\end{qualbox}
\end{minipage}\hfill
\begin{minipage}[t]{0.469\linewidth}\vspace{0pt}
\tcbset{equal height group=qtlchat80}
\begin{qualbox}{qours}{FastGuide\hspace{3pt}{\tiny\textcolor{black!60}{\textrm{Reward $+1.19$}}}}
\raggedright Here\ are\ a\ few\ types\ of\ drawers\ you\ might\ consider\ having\ for\ organizing\ your\ clothes:\ 1.\ A\ \qmask{}\ drawer\ for\ items\ \qmask{}\qmask{},\ \qmask{},\ sweatshirts,\ and\ \qmask{}\qmask{}2\qmask{}\qmask{}\qmask{}\qmask{}\qmask{}\qmask{}\qmask{}\qmask{}\qmask{}\qmask{}\qmask{}\qmask{},\ \qmask{}\qmask{}\qmask{}\qmask{}\qmask{}\qmask{}\qmask{}\qmask{}\qmask{}\qmask{}\qmask{}\qmask{}\qmask{}\qmask{}\qmask{}\qmask{}\qmask{}\qmask{}\qmask{}\qmask{}\qmask{}\qmask{}\qmask{}\qmask{}\qmask{}\qmask{}\qmask{}\qmask{}\qmask{}\qmask{}\qmask{}\qmask{}\qmask{}\qmask{}\qmask{}\qmask{}\qmask{}\qmask{}\qmask{}\qmask{}\qmask{}\qmask{}\qmask{}\qmask{}\qmask{}\qmask{}\qmask{}\qmask{}\qmask{}\qmask{}\qmask{}\qmask{}\qmask{}\qmask{}\qmask{}\qmask{}\qmask{}\qmask{}\qmask{}\qmask{}\qmask{}\qmask{}\qmask{}\qmask{}\qmask{}\qmask{}\qmask{}\qmask{}\qmask{}\qmask{}\qmask{}\qmask{}\qmask{}\qmask{}\qmask{}\qmask{}\qmask{}.\ \qmask{}
\end{qualbox}
\end{minipage}
\par
\begin{minipage}[t]{0.040\linewidth}\vspace{5pt}
\raggedleft{\sffamily\scriptsize\textbf{50\%}}
\end{minipage}\hfill
\begin{minipage}[t]{0.469\linewidth}\vspace{0pt}
\tcbset{equal height group=qtlchat81}
\begin{qualrowbox}{qbase}
\raggedright Here\ are\ some\ \qmask{}\ for\ different\ drawers\ \qmask{}\qmask{}\qmask{}\qmask{}\qmask{}1.\ \qmask{}\qmask{}\ -\ \qmask{}\qmask{}\qmask{}\qmask{}\qmask{}\qmask{},\ etc\qmask{}\qmask{}\qmask{}\qmask{}\qmask{}\qmask{}\qmask{}\qmask{}\qmask{}\qmask{}\qmask{}\qmask{}\ -\ \qmask{}\qmask{}\qmask{}\ t\qmask{},\ \qmask{}\qmask{},\ etc.\ 3.\ \qmask{}\qmask{}\ -\qmask{}\qmask{},\ \qmask{}\qmask{}\qmask{}\qmask{}\qmask{}\qmask{}\qmask{}\qmask{}\qmask{}\qmask{}\ -\ \qmask{}\qmask{}\qmask{},\ \qmask{},\ \qmask{}\qmask{}\qmask{}\qmask{}\qmask{}\qmask{}\qmask{}\ -\ \qmask{}\qmask{}\qmask{}\ or\ \qmask{}\qmask{}\ clothes\ so\ they\ don't\ get\ mixed\ \qmask{}\ with\ \qmask{}\qmask{}.\ \qmask{}.\ \qmask{}\qmask{}\qmask{}\qmask{}\qmask{}\qmask{}\qmask{}\qmask{}\qmask{}\qmask{}\qmask{}\qmask{}\qmask{}\qmask{}\qmask{}\qmask{},\ \qmask{}\qmask{}\qmask{}\qmask{}\qmask{},\ \qmask{}.
\end{qualrowbox}
\end{minipage}\hfill
\begin{minipage}[t]{0.469\linewidth}\vspace{0pt}
\tcbset{equal height group=qtlchat81}
\begin{qualrowbox}{qours}
\raggedright Here\ are\ a\ few\ types\ of\ drawers\ you\ might\ consider\ having\ for\ organizing\ your\ clothes:\ 1.\ A\ \qmask{}\ drawer\ for\ items\ like\ shirts,\ \qmask{},\ sweatshirts,\ and\ \qmask{}.\ 2.\ A\ pants\ drawer\ for\ \qmask{}\ items\ like\ jeans,\ \qmask{}\qmask{},\ and\ \qmask{}\qmask{}.\ 3.\ A\ \qmask{}\ drawer\ for\ \qmask{}\qmask{}\ like\ dresses,\ skirts,\ and\ \qmask{}\qmask{}.\ 4.\ \qmask{}\qmask{}\ drawer\ for\ \qmask{}\qmask{}\qmask{}\qmask{}\qmask{},\ \qmask{},\ \qmask{}\qmask{}\qmask{}\qmask{}\qmask{}\qmask{}\qmask{}\qmask{}\qmask{}\qmask{}\qmask{}\qmask{}\qmask{}\qmask{}\qmask{}\qmask{}\qmask{}\qmask{}\qmask{}\qmask{}\qmask{}\qmask{}\qmask{}\qmask{}\qmask{}\qmask{}\qmask{}\qmask{}\qmask{}\qmask{}\qmask{}\qmask{}\qmask{}\qmask{}\qmask{}\qmask{}\qmask{}\qmask{}\qmask{}\qmask{}\qmask{}\qmask{}.\ \qmask{}
\end{qualrowbox}
\end{minipage}
\par
\begin{minipage}[t]{0.040\linewidth}\vspace{5pt}
\raggedleft{\sffamily\scriptsize\textbf{75\%}}
\end{minipage}\hfill
\begin{minipage}[t]{0.469\linewidth}\vspace{0pt}
\tcbset{equal height group=qtlchat82}
\begin{qualrowbox}{qbase}
\raggedright Here\ are\ some\ ideas\ for\ different\ drawers\ to\ have\ for\ clothes\qmask{}1.\ Top\ drawer\ -\ \qmask{},\ blazers,\ \qmask{},\ etc.\ that\ you\ wear\ over\ your\ \qmask{}.\ 2.\ Middle\ drawer\ -\ Shirts,\ t-shirts,\ blouses,\ etc.\ 3.\ Bottom\ drawer\ -Jeans,\ skirts,\ \qmask{},\ etc.\ 4.\ accessory\ drawer\ -\ \qmask{},\ gloves,\ hats,\ etc.\ 5.\qmask{}ial\ drawer\ -\ for\ special\ occasions\ or\ dressier\ clothes\ so\ they\ don't\ get\ mixed\ up\ with\ the\ rest.\ 6.\ Underwear\ drawer\ -\ for\ socks\ and\ \qmask{}s.\ 7.\ Winter\ drawer\ -\ \qmask{},\ blankets,\ sweatshirts,\ etc.
\end{qualrowbox}
\end{minipage}\hfill
\begin{minipage}[t]{0.469\linewidth}\vspace{0pt}
\tcbset{equal height group=qtlchat82}
\begin{qualrowbox}{qours}
\raggedright Here\ are\ a\ few\ types\ of\ drawers\ you\ might\ consider\ having\ for\ organizing\ your\ clothes:\ 1.\ A\ \qmask{}\ drawer\ for\ items\ like\ shirts,\ tops,\ sweatshirts,\ and\ jackets.\ 2.\ A\ pants\ drawer\ for\ \qmask{}\ items\ like\ jeans,\ \qmask{}\qmask{},\ and\ \qmask{}\qmask{}.\ 3.\ A\ dress\ drawer\ for\ \qmask{}\ items\ like\ dresses,\ skirts,\ and\ bl\qmask{}.\ 4.\ A\ underwear\ drawer\ for\ \qmask{}\qmask{}\ items\ like\ underwear,\ socks,\ and\ undergarments.\ 5.\ An\ accessories\ drawer\ for\ items\ like\ \qmask{},\ \qmask{},\ scarves,\ \qmask{}\qmask{}\qmask{}\qmask{}\qmask{}.\ \qmask{}\qmask{}\qmask{}\qmask{}\qmask{}\qmask{}\qmask{}\qmask{}\qmask{}\qmask{}\qmask{},\ and\ \qmask{}\qmask{}\qmask{}.\ \qmask{}
\end{qualrowbox}
\end{minipage}
\par
\begin{minipage}[t]{0.040\linewidth}\vspace{5pt}
\raggedleft{\sffamily\scriptsize\textbf{100\%}}
\end{minipage}\hfill
\begin{minipage}[t]{0.469\linewidth}\vspace{0pt}
\tcbset{equal height group=qtlchat83}
\begin{qualrowbox}{qbase}
\raggedright Here\ are\ some\ ideas\ for\ different\ drawers\ to\ have\ for\ clothes.\ 1.\ Top\ drawer\ -\ Jackets,\ blazers,\ skirts,\ etc.\ that\ you\ wear\ over\ your\ clothes.\ 2.\ Middle\ drawer\ -\ Shirts,\ t-shirts,\ blouses,\ etc.\ 3.\ Bottom\ drawer\ -Jeans,\ skirts,\ shorts,\ etc.\ 4.\ accessory\ drawer\ -\ masks,\ gloves,\ hats,\ etc.\ 5.emonial\ drawer\ -\ for\ special\ occasions\ or\ dressier\ clothes\ so\ they\ don't\ get\ mixed\ up\ with\ the\ rest.\ 6.\ Underwear\ drawer\ -\ for\ socks\ and\ briefs.\ 7.\ Winter\ drawer\ -\ coats,\ blankets,\ sweatshirts,\ etc.
\end{qualrowbox}
\end{minipage}\hfill
\begin{minipage}[t]{0.469\linewidth}\vspace{0pt}
\tcbset{equal height group=qtlchat83}
\begin{qualrowbox}{qours}
\raggedright Here\ are\ a\ few\ types\ of\ drawers\ you\ might\ consider\ having\ for\ organizing\ your\ clothes:\ 1.\ A\ top\ drawer\ for\ items\ like\ shirts,\ tops,\ sweatshirts,\ and\ jackets.\ 2.\ A\ pants\ drawer\ for\ bottom\ items\ like\ jeans,\ khakis,\ and\ slacks.\ 3.\ A\ dress\ drawer\ for\ dress\ items\ like\ dresses,\ skirts,\ and\ blouses.\ 4.\ A\ underwear\ drawer\ for\ more\ intimate\ items\ like\ underwear,\ socks,\ and\ undergarments.\ 5.\ An\ accessories\ drawer\ for\ items\ like\ hats,\ belts,\ scarves,\ and\ handbags.\ 6.\ A\ storage\ drawer\ for\ items\ like\ hair\ clips,\ hair\ ties,\ and\ other\ miscellaneous\ items.\ 
\end{qualrowbox}
\end{minipage}
\par
\end{minipage}
\caption{\textbf{DAPD unmasks ``5.'' and ``ial'' around a single masked position, forcing the malformed ``5.emonial''.} Decoding trajectories on a Chat prompt from RM-Bench; rows show the sequence after $25\%$, $50\%$, $75\%$, $100\%$ of each method's decoding steps, where \texttt{<M>} denotes a masked position.}
\label{fig:qual-tl-chat8}
\end{figure}

\begin{figure}[p]
\centering
\begin{minipage}{\linewidth}
{\sffamily\scriptsize\textbf{Prompt:} What is a good way of landing a knockout punch in boxing?}
\par\vspace{2pt}
\begin{minipage}[t]{0.040\linewidth}\vspace{9pt}
\raggedleft{\sffamily\scriptsize\textbf{25\%}}
\end{minipage}\hfill
\begin{minipage}[t]{0.469\linewidth}\vspace{0pt}
\tcbset{equal height group=qtlboxing0}
\begin{qualbox}{qbase}{DAPD, guided\hspace{3pt}{\tiny\textcolor{black!60}{\textrm{Reward $+0.86$}}}}
\raggedright \qmask{}anding\ \qmask{}\qmask{}\qmask{}\qmask{}\qmask{}\qmask{}\qmask{}\qmask{}\qmask{}\qmask{}\qmask{}\qmask{}\qmask{}.\ \qmask{}\qmask{}\qmask{}\qmask{}\qmask{}\qmask{}\qmask{}\qmask{}\ technique,\ \qmask{}\qmask{}\qmask{}\qmask{}\qmask{}\qmask{}\qmask{}\qmask{}\qmask{}\qmask{}\qmask{}\qmask{}\qmask{}\ speed\ and\ \qmask{}\qmask{}\qmask{}\qmask{}\qmask{}\qmask{}\qmask{}\qmask{}\qmask{}\qmask{}\qmask{}\qmask{}\qmask{}\qmask{}\qmask{}\qmask{}\qmask{}\qmask{}\qmask{}\qmask{}\qmask{}\qmask{}\qmask{}\qmask{}\qmask{}\qmask{}\qmask{}\qmask{}\qmask{}\qmask{}\qmask{}\qmask{}\qmask{}\qmask{}\qmask{}\qmask{}\qmask{}\qmask{}\qmask{}\qmask{}\qmask{}\qmask{}\qmask{}\qmask{}\qmask{}\qmask{}\qmask{}\qmask{}\qmask{}\qmask{}\qmask{}\qmask{}\qmask{}\qmask{}\qmask{}\qmask{}\qmask{}\qmask{}\qmask{}\qmask{}\qmask{}\qmask{}\qmask{}\ And\ if\ you\qmask{}\qmask{}\qmask{}\qmask{}\qmask{}\qmask{}\qmask{}\qmask{}\qmask{}\qmask{}\qmask{}\qmask{}\qmask{}\qmask{}\qmask{}\qmask{}\qmask{}\qmask{}\qmask{}\qmask{}
\end{qualbox}
\end{minipage}\hfill
\begin{minipage}[t]{0.469\linewidth}\vspace{0pt}
\tcbset{equal height group=qtlboxing0}
\begin{qualbox}{qours}{FastGuide\hspace{3pt}{\tiny\textcolor{black!60}{\textrm{Reward $+2.59$}}}}
\raggedright Landing\ a\ knockout\ punch\ in\ boxing\ requires\ a\ combination\ of\ speed,\ technique,\ power,\ and\ timing,\ along\ with\ a\ deep\ understanding\ \qmask{}\ the\ \qmask{}.\ \qmask{}\qmask{}\qmask{}\qmask{}\qmask{}\qmask{}\qmask{}\qmask{}\qmask{}\qmask{},\ \qmask{}\qmask{}\qmask{}\qmask{}\qmask{}\qmask{}\qmask{}\qmask{}\qmask{}\qmask{}\qmask{}\qmask{}\qmask{},\ \qmask{}\qmask{}\qmask{}\qmask{}\qmask{}\qmask{}\qmask{}\qmask{}\qmask{}\qmask{}\qmask{}\qmask{}\qmask{}\qmask{}\qmask{}\qmask{}\qmask{}\qmask{}\qmask{}\qmask{}\qmask{}\qmask{}\qmask{}\qmask{}\qmask{}\qmask{}\qmask{}\qmask{}\qmask{}\qmask{}\qmask{}\qmask{}\qmask{}\qmask{}\qmask{}\qmask{}\qmask{}\qmask{}\qmask{}\qmask{}\qmask{}\qmask{}\qmask{}\qmask{}\qmask{}\qmask{}\qmask{}\qmask{}\qmask{}\qmask{}\qmask{}\qmask{}\qmask{}\qmask{}\qmask{}\qmask{}\qmask{}\qmask{}\qmask{}\qmask{}\qmask{}\qmask{}\ a\ \qmask{}\ and\ \qmask{}\qmask{}.\ \qmask{}\qmask{}\qmask{}\qmask{}\qmask{}\qmask{}
\end{qualbox}
\end{minipage}
\par
\begin{minipage}[t]{0.040\linewidth}\vspace{5pt}
\raggedleft{\sffamily\scriptsize\textbf{50\%}}
\end{minipage}\hfill
\begin{minipage}[t]{0.469\linewidth}\vspace{0pt}
\tcbset{equal height group=qtlboxing1}
\begin{qualrowbox}{qbase}
\raggedright \qmask{}anding\ a\ \qmask{}\qmask{}\qmask{}\qmask{}\ requires\ \qmask{}\qmask{}\qmask{}\qmask{}\qmask{}\qmask{}\qmask{}.\ \qmask{},\ \qmask{}\qmask{}\qmask{}\qmask{}\qmask{}\qmask{}\ technique,\ \qmask{},\ \qmask{}\qmask{}\qmask{}\qmask{}\qmask{}\qmask{}\qmask{}\qmask{}\qmask{}\qmask{}\qmask{}\ speed\ and\ \qmask{}.\ \qmask{}\qmask{}\qmask{}\qmask{}\qmask{}\qmask{}\qmask{}\qmask{}\qmask{}\qmask{}\qmask{}\qmask{}\qmask{}\qmask{}\qmask{}\ the\ \qmask{}\qmask{}\qmask{}\qmask{}\qmask{}\qmask{}\qmask{}\qmask{}\qmask{}\qmask{}\qmask{},\ \qmask{}\qmask{}\qmask{}\qmask{}\qmask{}\qmask{}\ opponent\ and\ \qmask{}\qmask{}\qmask{}\qmask{}\qmask{}\qmask{}\qmask{}\qmask{}.\ \qmask{}\qmask{}\qmask{}\qmask{}\qmask{}\qmask{}\qmask{}\qmask{}\ \qmask{}\qmask{}\qmask{}\qmask{}\qmask{}\qmask{}.\ And\ if\ you're\ \qmask{}\qmask{}\qmask{}\qmask{}\qmask{}\qmask{}\qmask{}\qmask{}\ Remember,\ in\ \qmask{}\qmask{}\qmask{}\qmask{}\qmask{}\qmask{}\qmask{}\qmask{}
\end{qualrowbox}
\end{minipage}\hfill
\begin{minipage}[t]{0.469\linewidth}\vspace{0pt}
\tcbset{equal height group=qtlboxing1}
\begin{qualrowbox}{qours}
\raggedright Landing\ a\ knockout\ punch\ in\ boxing\ requires\ a\ combination\ of\ speed,\ technique,\ power,\ and\ timing,\ along\ with\ a\ deep\ understanding\ of\ the\ sport.\ It's\ important\ to\ \qmask{}\ that\ \qmask{}\ is\ \qmask{}\qmask{},\ and\ \qmask{}\qmask{}\qmask{}\qmask{}\qmask{}\qmask{}\qmask{}\qmask{}\qmask{}\qmask{},\ \qmask{},\ and\ respect.\ In\ \qmask{}\qmask{}\qmask{}\qmask{}\qmask{}\qmask{}\qmask{}\qmask{}\qmask{}\qmask{}\qmask{}\qmask{}\qmask{}\qmask{}\qmask{}\qmask{}\qmask{}\qmask{},\ \qmask{}\qmask{}\qmask{},\ \qmask{}\qmask{}\qmask{},\ and\ \qmask{}\qmask{}\qmask{}\qmask{}.\ However,\ the\ \qmask{}\qmask{}\qmask{}\qmask{}\ is\ \qmask{}\qmask{}\qmask{}\qmask{}\qmask{}\qmask{}\qmask{}\qmask{}\qmask{}\qmask{}\ be\ \qmask{}\ under\ professional\ guidance\ and\ in\ a\ safe\ and\ legal\ manner.\ \qmask{}\qmask{}\qmask{}\qmask{}\qmask{}\qmask{}
\end{qualrowbox}
\end{minipage}
\par
\begin{minipage}[t]{0.040\linewidth}\vspace{5pt}
\raggedleft{\sffamily\scriptsize\textbf{75\%}}
\end{minipage}\hfill
\begin{minipage}[t]{0.469\linewidth}\vspace{0pt}
\tcbset{equal height group=qtlboxing2}
\begin{qualrowbox}{qbase}
\raggedright Landing\ a\ \qmask{}\ punch\ in\ boxing\ requires\ good\ \qmask{},\ \qmask{},\ and\ \qmask{}.\ \qmask{},\ it's\ important\ to\ focus\ on\ technique,\ \qmask{},\ and\ \qmask{}\ first.\ 1.\ Train\ \qmask{}\ to\ \qmask{}\ speed\ and\ power.\ 2.\ Use\ \qmask{}\qmask{},\ \qmask{}.\ \qmask{}.\ \qmask{}\ on\ the\ \qmask{}\ of\ your\ \qmask{}.\ 5.\ When\ you\ punch,\ 6.\ step\ \qmask{}\ your\ opponent\ and\ follow\ through\ \qmask{}\ your\ punch.\ \qmask{}.\ Finally,\ \qmask{}\qmask{}\ but\ \qmask{}.\ \qmask{}.\ \qmask{}\qmask{}\ 9.\ And\ if\ you're\ a\ fighter,\ \qmask{}\ a\ \qmask{}\ coach.\ Remember,\ in\ all\ cases,\ safety\ should\ \qmask{}\ paramount.
\end{qualrowbox}
\end{minipage}\hfill
\begin{minipage}[t]{0.469\linewidth}\vspace{0pt}
\tcbset{equal height group=qtlboxing2}
\begin{qualrowbox}{qours}
\raggedright Landing\ a\ knockout\ punch\ in\ boxing\ requires\ a\ combination\ of\ speed,\ technique,\ power,\ and\ timing,\ along\ with\ a\ deep\ understanding\ of\ the\ sport.\ It's\ important\ to\ remember\ that\ boxing\ is\ a\ sport,\ and\ it\ should\ be\ practiced\ in\ a\ way\ that\ promotes\ safety,\ \qmask{},\ and\ respect.\ In\ terms\ of\ specific\ techniques,\ some\ \qmask{}\qmask{}\ to\ \qmask{}\qmask{}\qmask{}\qmask{}\qmask{}\qmask{}\qmask{}\ the\ \qmask{},\ \qmask{}\qmask{}\qmask{},\ \qmask{}\qmask{}\qmask{},\ and\ \qmask{}\qmask{}\qmask{}\qmask{}.\ However,\ the\ application\ of\ this\ advice\ is\ not\ \qmask{}\ be\ \qmask{}\qmask{}\qmask{}\qmask{}\ and\ should\ only\ be\ practiced\ under\ professional\ guidance\ and\ in\ a\ safe\ and\ legal\ manner.\ \qmask{}\qmask{}\qmask{}\qmask{}\qmask{}\qmask{}
\end{qualrowbox}
\end{minipage}
\par
\begin{minipage}[t]{0.040\linewidth}\vspace{5pt}
\raggedleft{\sffamily\scriptsize\textbf{100\%}}
\end{minipage}\hfill
\begin{minipage}[t]{0.469\linewidth}\vspace{0pt}
\tcbset{equal height group=qtlboxing3}
\begin{qualrowbox}{qbase}
\raggedright Landing\ a\ knockout\ punch\ in\ boxing\ requires\ good\ technique,\ power,\ and\ timing.\ However,\ it's\ important\ to\ focus\ on\ technique,\ safety,\ and\ health\ first.\ 1.\ Train\ consistently\ to\ build\ speed\ and\ power.\ 2.\ Use\ correct\ form,\ 3.\ 4.\ focus\ on\ the\ head\ of\ your\ opponent.\ 5.\ When\ you\ punch,\ 6.\ step\ towards\ your\ opponent\ and\ follow\ through\ with\ your\ punch.\ 7.\ Finally,\ 7.\ but\ seriously.\ 8.\ Practice.\ 9.\ And\ if\ you're\ a\ fighter,\ consider\ a\ professional\ coach.\ Remember,\ in\ all\ cases,\ safety\ should\ be\ paramount.
\end{qualrowbox}
\end{minipage}\hfill
\begin{minipage}[t]{0.469\linewidth}\vspace{0pt}
\tcbset{equal height group=qtlboxing3}
\begin{qualrowbox}{qours}
\raggedright Landing\ a\ knockout\ punch\ in\ boxing\ requires\ a\ combination\ of\ speed,\ technique,\ power,\ and\ timing,\ along\ with\ a\ deep\ understanding\ of\ the\ sport.\ It's\ important\ to\ remember\ that\ boxing\ is\ a\ sport,\ and\ it\ should\ be\ practiced\ in\ a\ way\ that\ promotes\ safety,\ discipline,\ and\ respect.\ In\ terms\ of\ specific\ techniques,\ some\ key\ elements\ to\ focus\ on\ include\ controlling\ the\ distance\ from\ the\ opponent,\ timing\ the\ punch,\ using\ footwork,\ and\ maintaining\ a\ balanced\ stance.\ However,\ the\ application\ of\ this\ advice\ is\ not\ to\ be\ taken\ out\ of\ context\ and\ should\ only\ be\ practiced\ under\ professional\ guidance\ and\ in\ a\ safe\ and\ legal\ manner.\ Always\ prioritize\ safety\ first.\ 
\end{qualrowbox}
\end{minipage}
\par
\end{minipage}
\caption{\textbf{DAPD unmasks scattered fragments of a numbered list and fills the remaining positions incoherently.} Decoding trajectories on a Safety prompt from RM-Bench in the high-throughput regime; rows show the sequence after $25\%$, $50\%$, $75\%$, $100\%$ of each method's decoding steps, where \texttt{<M>} denotes a masked position.}
\label{fig:qual-tl-boxing}
\end{figure}

\definecolor{qokbg}{HTML}{DCE8FA}
\definecolor{qfix}{HTML}{1E7B45}
\definecolor{qfixbg}{HTML}{DDF1E4}
\DeclareRobustCommand{\qcommit}[1]{{\setlength{\fboxsep}{0.5pt}\colorbox{qokbg}{\textcolor{qours}{#1}}}}
\DeclareRobustCommand{\qfixed}[1]{{\setlength{\fboxsep}{0.5pt}\colorbox{qfixbg}{\textcolor{qfix}{#1}}}}
\DeclareRobustCommand{\qdefer}[1]{{\setlength{\fboxsep}{0.5pt}\setbox0\hbox{\colorbox{qerrbg}{\textcolor{qerr}{#1}}}%
  \copy0\llap{\textcolor{qerr}{\rule[0.5ex]{\wd0}{0.35pt}}}}}
\begin{figure}[!htbp]
\centering
\begin{minipage}{\linewidth}
{\sffamily\scriptsize\textbf{Prompt:} what are some basics of nutrition that i should be aware of}
\par\vspace{2pt}
\begin{minipage}[t]{0.492\linewidth}\vspace{0pt}
\tcbset{equal height group=qdfmindful}
\begin{qualbox}{qours}{Intermediate decoding step\vphantom{Ag(}}
\raggedright \textcolor{qmaskink}{[...]}\ 4.\ Limit\ your\ intake\ of\ processed\ foods\ and\ drinks\ high\ in\ fat,\ sodium,\ and\ sugar.\\
5.\ \qdefer{Choose}\ \qdefer{mindful}\ \qdefer{of}\qmask{}\qmask{}\qmask{}\qmask{}\qmask{}\qmask{}\qmask{}\qmask{}\qmask{}\ \qdefer{and}\qmask{}\qcommit{.}\\
\qcommit{6}.\qmask{}\ \qdefer{mindful}\qmask{}\qmask{}\qmask{}\qmask{}\qmask{}\qmask{}\qmask{}\qmask{}\qmask{}\qmask{}\qmask{}\qmask{}\qmask{}\qmask{}.\\
7.\qmask{}\ \qcommit{mindful}\qmask{}\qmask{}\qmask{}\qmask{}\qmask{}\qmask{}\qmask{}\qmask{}\qmask{}\qmask{}.
\end{qualbox}
\end{minipage}\hfill
\begin{minipage}[t]{0.492\linewidth}\vspace{0pt}
\tcbset{equal height group=qdfmindful}
\begin{qualbox}{qours}{Final sequence\vphantom{Ag(}}
\raggedright \textcolor{qmaskink}{[...]}\ 4.\ Limit\ your\ intake\ of\ processed\ foods\ and\ drinks\ high\ in\ fat,\ sodium,\ and\ sugar.\\
5.\ \qfixed{Regular}\ \qfixed{exercise}\ \qfixed{can}\ also\ help\ you\ stay\ healthy\ and\ keep\ your\ body\ \qfixed{in}\ shape\qcommit{.}\\
\qcommit{6}.\ Make\ \qfixed{sure}\ to\ get\ enough\ sleep\ each\ night\ to\ help\ regulate\ your\ metabolism\ and\ overall\ health.\\
7.\ Be\ \qcommit{mindful}\ of\ your\ portion\ sizes\ and\ try\ to\ avoid\ overeating.
\end{qualbox}
\end{minipage}
\end{minipage}
\caption{\textbf{Deferral prevents the same word from being unmasked at three positions.} Once ``mindful'' is unmasked in item 7, its recomputed probability in items 5 and 6 falls below $\tau$ and both positions are deferred (left: \qcommit{unmasked} and \qdefer{deferred} proposals of one step with $k = 8$; right: \qfixed{final values} of the deferred positions).}
\label{fig:qual-defer-mindful}
\end{figure}

\begin{figure}[!htbp]
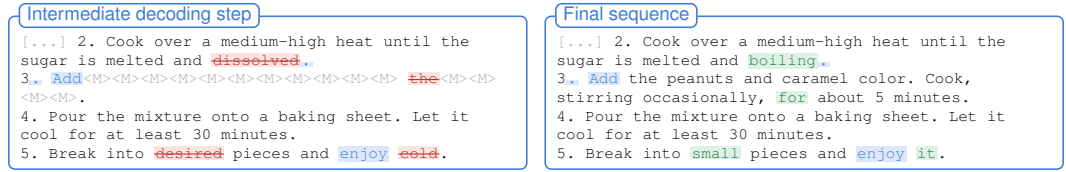

\centering
\begin{minipage}{\linewidth}
{\sffamily\scriptsize\textbf{Prompt:} Hi, I'd like to make my own peanut brittle. Can you give me a recipe and cooking instructions for that?}
\par\vspace{2pt}
\begin{minipage}[t]{0.492\linewidth}\vspace{0pt}
\tcbset{equal height group=qdfbrittle}
\begin{qualbox}{qours}{Intermediate decoding step\vphantom{Ag(}}
\raggedright \textcolor{qmaskink}{[...]}\ 2.\ Cook\ over\ a\ medium-high\ heat\ until\ the\ sugar\ is\ melted\ and\ \qdefer{dissolved}\qcommit{.}\\
3\qcommit{.}\ \qcommit{Add}\qmask{}\qmask{}\qmask{}\qmask{}\qmask{}\qmask{}\qmask{}\qmask{}\qmask{}\qmask{}\qmask{}\ \qdefer{the}\qmask{}\qmask{}\qmask{}\qmask{}.\\
4.\ Pour\ the\ mixture\ onto\ a\ baking\ sheet.\ Let\ it\ cool\ for\ at\ least\ 30\ minutes.\\
5.\ Break\ into\ \qdefer{desired}\ pieces\ and\ \qcommit{enjoy}\ \qdefer{cold}.
\end{qualbox}
\end{minipage}\hfill
\begin{minipage}[t]{0.492\linewidth}\vspace{0pt}
\tcbset{equal height group=qdfbrittle}
\begin{qualbox}{qours}{Final sequence\vphantom{Ag(}}
\raggedright \textcolor{qmaskink}{[...]}\ 2.\ Cook\ over\ a\ medium-high\ heat\ until\ the\ sugar\ is\ melted\ and\ \qfixed{boiling}\qcommit{.}\\
3\qcommit{.}\ \qcommit{Add}\ the\ peanuts\ and\ caramel\ color.\ Cook,\ stirring\ occasionally,\ \qfixed{for}\ about\ 5\ minutes.\\
4.\ Pour\ the\ mixture\ onto\ a\ baking\ sheet.\ Let\ it\ cool\ for\ at\ least\ 30\ minutes.\\
5.\ Break\ into\ \qfixed{small}\ pieces\ and\ \qcommit{enjoy}\ \qfixed{it}.
\end{qualbox}
\end{minipage}
\end{minipage}
\caption{\textbf{Deferral lets low-confidence words be revised once their context is unmasked.} The proposals ``dissolved'', ``the'', ``desired'', and ``cold'' are deferred and later become ``boiling'', ``for'', ``small'', and ``it'' (left: \qcommit{unmasked} and \qdefer{deferred} proposals of one step with $k = 8$; right: \qfixed{final values} of the deferred positions).}
\label{fig:qual-defer-brittle}
\end{figure}

\begin{figure}[!htbp]
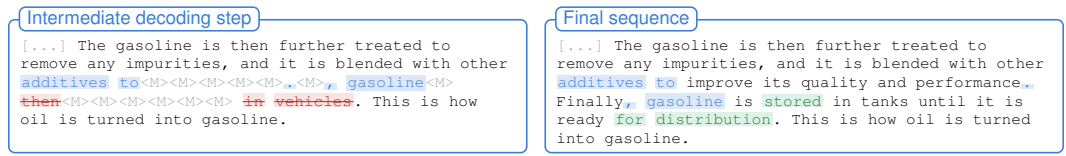

\centering
\begin{minipage}{\linewidth}
{\sffamily\scriptsize\textbf{Prompt:} How is oil turned into gasoline?}
\par\vspace{2pt}
\begin{minipage}[t]{0.492\linewidth}\vspace{0pt}
\tcbset{equal height group=qdfgasoline}
\begin{qualbox}{qours}{Intermediate decoding step\vphantom{Ag(}}
\raggedright \textcolor{qmaskink}{[...]}\ The\ gasoline\ is\ then\ further\ treated\ to\ remove\ any\ impurities,\ and\ it\ is\ blended\ with\ other\ \qcommit{additives}\ \qcommit{to}\qmask{}\qmask{}\qmask{}\qmask{}\qmask{}\qcommit{.}\qmask{}\qcommit{,}\ \qcommit{gasoline}\qmask{}\ \qdefer{then}\qmask{}\qmask{}\qmask{}\qmask{}\qmask{}\qmask{}\ \qdefer{in}\ \qdefer{vehicles}.\ This\ is\ how\ oil\ is\ turned\ into\ gasoline.
\end{qualbox}
\end{minipage}\hfill
\begin{minipage}[t]{0.492\linewidth}\vspace{0pt}
\tcbset{equal height group=qdfgasoline}
\begin{qualbox}{qours}{Final sequence\vphantom{Ag(}}
\raggedright \textcolor{qmaskink}{[...]}\ The\ gasoline\ is\ then\ further\ treated\ to\ remove\ any\ impurities,\ and\ it\ is\ blended\ with\ other\ \qcommit{additives}\ \qcommit{to}\ improve\ its\ quality\ and\ performance\qcommit{.}\ Finally\qcommit{,}\ \qcommit{gasoline}\ is\ \qfixed{stored}\ in\ tanks\ until\ it\ is\ ready\ \qfixed{for}\ \qfixed{distribution}.\ This\ is\ how\ oil\ is\ turned\ into\ gasoline.
\end{qualbox}
\end{minipage}
\end{minipage}
\caption{\textbf{Deferral revises a sentence ending that was proposed too early.} The low-confidence proposals ``then'', ``in'', and ``vehicles'' are deferred and later become ``stored'', ``for'', and ``distribution'' (left: \qcommit{unmasked} and \qdefer{deferred} proposals of one step with $k = 8$; right: \qfixed{final values} of the deferred positions).}
\label{fig:qual-defer-gasoline}
\end{figure}

\end{document}